\documentclass[12pt]{article}
\usepackage{graphicx}
\usepackage{amssymb}
\usepackage{amsmath}
\usepackage{amsbsy}
\usepackage{latexsym}
\usepackage{subfigure}
\usepackage{lscape}
\usepackage{multirow}
\usepackage{bm}
\usepackage{url}

\newcommand{\argmin}{\mathop{\mathrm{argmin}}}
\newcommand{\argmax}{\mathop{\mathrm{argmax}}}
\renewcommand{\max}{\mathop{\mathrm{max}}}

\newcommand{\trans}[1]{#1^\top}
\newcommand{\unorm}[1]{\bigl\|#1\bigr\|}

\newcommand{\unorms}[1]{\unorm{#1}^2}

\newcommand{\hpara}{\boldsymbol{\rho}}
\newcommand{\hmean}{\boldsymbol{\eta}}
\newcommand{\hvar}{\boldsymbol{\tau}}
\newcommand{\para}{\boldsymbol{\theta}}
\newcommand{\pmean}{\boldsymbol{\mu}}
\newcommand{\pstd}{\sigma}
\newcommand{\paran}[1]{\para_{#1}}
\newcommand{\optpara}{\para^{*}}
\newcommand{\A}{{\mathcal A}}
\renewcommand{\S}{{\mathcal S}}

\newcommand{\pIs}[1]{p(#1)}

\newcommand{\psa}[3]{p(#1|#2,#3)}
\newcommand{\qsa}[3]{q(#1|#2,#3)}
\newcommand{\pa}[3]{p(#1|#2,#3)}
\newcommand{\ph}[2]{p(#1|#2)}
\newcommand{\R}{R}
\newcommand{\Rsa}[3]{\R(#1,#2,#3)}
\renewcommand{\r}{r}
\newcommand{\rh}[1]{\r(#1)}

\newcommand{\s}{\boldsymbol{s}}
\newcommand{\st}[1]{\s_{#1}}
\newcommand{\snt}[2]{\s^{#1}_{#2}}
\renewcommand{\a}{a}
\newcommand{\at}[1]{\a_{#1}}
\newcommand{\ant}[2]{\a^{#1}_{#2}}
\newcommand{\h}{h}
\newcommand{\hn}[1]{\h_{#1}}

\newcommand{\Nstep}{T}
\newcommand{\nstep}{t}

\newcommand{\Nepi}{N}
\newcommand{\nepi}{n}

\newcommand{\rdis}{\gamma}
\newcommand{\J}[1]{J(#1)}
\newcommand{\grad}[1]{\nabla_{#1}}
\newcommand{\gradp}{\nabla_{\para}}
\newcommand{\Jhat}[1]{\widehat{J}(#1)}
\newcommand{\lrate}{\varepsilon}

\newcommand{\pdist}[2]{p(#1|#2)}
\newcommand{\condens}[2]{r(x,y)}

\newcommand{\G}[1]{G(#1)}
\newcommand{\Ghat}[1]{\widehat{G}(#1)}
\newcommand{\Gz}[1]{G_0(#1)}
\newcommand{\cbasismat}{\boldsymbol{H}}
\newcommand{\cbasisvec}{\boldsymbol{h}}
\newcommand{\cbasisbar}{\overline{\boldsymbol{\Phi}}}

\newcommand{\cbasismathat}{\widehat{\cbasismat}}
\newcommand{\cbasisvechat}{\widehat{\cbasisvec}}
\newcommand{\cparam}{\boldsymbol{\alpha}}
\newcommand{\cparamtilde}{\widetilde{\cparam}}
\newcommand{\cparamhat}{\widehat{\cparam}}
\newcommand{\cbasis}{\boldsymbol{\phi}}
\newcommand{\I}[1]{\boldsymbol{I}_{#1}}

\newcommand{\phat}[1]{\widehat{p}(#1)}

\newcommand{\Jhatb}[2]{\widehat{J}^{#2}(#1)}
\newcommand{\bline}{b}

\newcommand{\Jhatbw}[3]{\widehat{J}_{#1}^{#2}(#3)}

\title{
\vspace*{-22mm}
Model-Based Policy Gradients\\
with Parameter-Based Exploration\\
by Least-Squares Conditional Density Estimation
}

\author{
Syogo Mori$^1$, Voot Tangkaratt$^1$, Tingting Zhao$^1$\\
Jun Morimoto$^2$, and Masashi Sugiyama$^1$\\[2mm]
$^1$Tokyo Institute of Technology, Japan.\\
\{mori@sg., voot@sg., tingting@sg., sugi@\}cs.titech.ac.jp\\[2mm]
$^2$ATR Computational Neuroscience Labs, Japan\\
xmorimo@atr.jp
}

\date{}
\begin{document}
\maketitle

\begin{abstract}
\noindent
The goal of reinforcement learning (RL) is to
let an agent learn an optimal control policy in an unknown environment
so that future expected rewards are maximized.
The model-free RL approach directly learns the policy based on data samples.
Although using many samples tends to improve the accuracy of policy learning,
collecting a large number of samples is often expensive in practice.
On the other hand, the model-based RL approach first estimates the transition model of the environment and then learns the policy based on the estimated transition model.
Thus, if the transition model is accurately learned from a small amount of data,
the model-based approach can perform better than the model-free approach.
In this paper, we propose a novel model-based RL method by combining a recently proposed
model-free policy search method called \emph{policy gradients with parameter-based exploration} and the state-of-the-art transition model estimator called \emph{least-squares conditional density estimation}.
Through experiments, we demonstrate the practical usefulness
of the proposed method.

\end{abstract}

\pagestyle{myheadings}
\markright{}
\sloppy

\section{Introduction}
\label{sec:introduction}

\emph{Reinforcement learning} (RL) is a framework to let
an agent learn an optimal control policy in an unknown environment
so that expected future rewards are maximized
\cite{JAIR:Kaelbling+Littman+Moore:1996}.
The RL methods developed so far can be categorized into two types:
\emph{Policy iteration}
where policies are learned based on value function approximation
\cite{book:Sutton+Barto:1998,JMLR:Lagoudakis+Parr:2003}
and \emph{policy search}
where policies are learned directly to maximize expected future rewards
\cite{mach:Williams:1992,nc:Dayan+Hinton:1997,%
SutMcASinMan00,nips02-CN11,NN:Sehnke+etal:2010,NC:Zhao+etal:2013}.

\subsection{Policy Iteration Vs.~Policy Search}
A value function represents expected future rewards as a function
of a state or a state and an action.
In the policy iteration framework,
approximation of the value function for the current policy
and
improvement of the policy based on the learned value function
are iteratively
performed until an optimal policy is found.
Thus, accurately approximating the value function is a challenge
in the value function based approach.
So far, various machine learning techniques have been
employed for better value function approximation,
such as least-squares approximation \cite{JMLR:Lagoudakis+Parr:2003},
manifold learning \cite{AutoRobo:Sugiyama+etal:2008},
efficient sample reuse \cite{NN:Hachiya+etal:2009},
active learning \cite{NN:Akiyama+etal:2010},
and
robust learning \cite{IEICE:Sugiyama+etal:2010b}.

However, because policy functions are learned indirectly via value functions
in policy iteration,
improving the quality of value function approximation does not necessarily
yield a better policy function.
Furthermore, because a small change in value functions
can cause a big change in policy functions,
it is not safe to use the value function based approach
for controlling expensive dynamic systems such as a humanoid robot.
Another weakness of the value function approach is that
it is difficult to handle continuous actions
because a maximizer of the value function with respect to an action
needs to be found for policy improvement.

On the other hand, in the policy search approach,
policy functions are determined so that
expected future rewards are directly maximized.
A popular policy search method is to update policy functions via gradient ascent.
However, a classic policy gradient method called REINFORCE \cite{mach:Williams:1992}
tends to produce gradient estimates with large variance,
which results in unreliable policy improvement
\cite{IROS:Peters+Schaal:2006}.
More theoretically, it was shown that the variance of policy gradients
can be proportional to the length of an agent's trajectory,
due to the stochasticity of policies \cite{NN:Zhao+etal:2012}.
This can be a critical limitation in RL problems with long trajectories.

To cope with this problem, a novel policy gradient method called
\emph{policy gradients with parameter-based exploration} (PGPE) was proposed
\cite{NN:Sehnke+etal:2010}.
In PGPE, deterministic policies are used to suppress irrelevant randomness
and useful stochasticity is introduced by drawing policy parameters
from a prior distribution.
Then, instead of policy parameters, hyper-parameters included in the prior distribution
are learned from data.
Thanks to this prior-based formulation, the variance of gradient estimates
in PGPE is independent of the length of an agent's trajectory \cite{NN:Zhao+etal:2012}.
However, PGPE still suffers from an instability problem in small sample cases.
To further improve the practical performance of PGPE,
an efficient sample reuse method called
\emph{importance-weighted PGPE} (IW-PGPE) was proposed recently
and demonstrated to achieve the state-of-the-art performance \cite{NC:Zhao+etal:2013}.

\subsection{Model-Based Vs.~Model-Free}
The RL methods reviewed above are categorized into the \emph{model-free} approach,
where policies are learned without explicitly modeling
the unknown environment (i.e., the transition probability of the agent in the environment).
On the other hand, an alternative approach called
the \emph{model-based} approach explicitly models the environment in advance
and uses the learned environment model for policy learning  \cite{Wang:2003,Deisenroth:2011}.
In the model-based approach,
no additional sampling cost is necessary to generate artificial samples from the learned
environment model.

The model-based approach is particularly advantageous in the policy search scenario.
For example, given a fixed budget for data collection,
IW-PGPE requires us to determine the \emph{sampling schedule} in advance.
More specifically, we need to decide, e.g.,
whether we gather many samples in the beginning
or only a small batch of samples are collected for a longer period.
However, optimizing the sampling schedule in advance is not possible
without strong prior knowledge. Thus, we need to just blindly
design the sampling schedule in practice, which can cause
significant performance degradation.
On the other hand, the model-based approach does not suffer from
this problem because
we can draw as many trajectory samples as we want
from the learned transition model without additional sampling costs.

Another advantage of the model-based approach lies in \emph{baseline subtraction}.
In the gradient-based policy search methods such as REINFORCE and PGPE,
subtraction of a baseline from a gradient estimate is a vital
technique to reduce the estimation variance
of policy gradients \cite{IROS:Peters+Schaal:2006,NC:Zhao+etal:2013}.
If the baseline is estimated from samples that are statistically independent
of samples used for the estimation of policy gradients,
variance reduction can be carried out without increasing the estimation bias.
However, such independent samples are not available in practice
(if available, they should be used for policy gradient estimation),
and thus variance reduction by baseline subtraction is practically
performed at the expense of bias increase.
On the other hand, in the model-based scenario,
we can draw as many trajectory samples as we want
from the learned transition model without additional sampling costs.
Therefore, two statistically independent sets of samples can be generated
and they can be separately used for policy gradient estimation and baseline estimation.

\subsection{Transition Model Learning by Least-Squares Conditional Density Estimation}

If the unknown environment is accurately approximated,
the model-based approach can fully enjoy all the above advantages.
However, accurately estimating the transition model from a limited amount of trajectory data
in multi-dimensional continuous state and action spaces is highly challenging.
Although the model-based method that does not require an accurate transition model
was developed \cite{Abbeel:2006},
it is only applicable to deterministic environments,
which significantly limits its range of applications in practice.
On the other hand, a recently proposed model-based policy search method called PILCO
\cite{Deisenroth:2011}
learns a probabilistic transition model by the Gaussian process (GP) \cite{book:Rasmussen+Williams:2006},
and explicitly incorporates long-term model uncertainty.
However, PILCO requires states and actions to follow Gaussian distributions
and the reward function to be a particular exponential form
to ensure that the policy evaluation is performed in a closed form
and policy gradients are computed analytically for policy improvement.
These strong requirements make PILCO practically restrictive.

To overcome such limitations of existing approaches,
we propose a highly practical policy-search algorithm
by extending the model-free PGPE method to the model-based scenario.
In the proposed model-based PGPE (M-PGPE) method,
the transition model is learned by the state-of-the-art
non-parametric conditional density estimator called
\emph{least-squares conditional density estimation} (LSCDE)
\cite{IEICE:Sugiyama+etal:2010a},
which has various superior properties:
It can directly handle multi-dimensional inputs and outputs,
it was proved to achieve the optimal convergence rate \cite{ML:Kanamori+etal:2012},
it has high numerical stability \cite{ML:Kanamori+etal:2012b},
it is robust against outliers \cite{AISM:Sugiyama+etal:2012},
its solution can be analytically and efficiently computed just by
solving a system of linear equations \cite{JMLR:Kanamori+etal:2009},
and generating samples from the learned conditional density is straightforward.
Through experiments,
we demonstrate that the proposed M-PGPE method is a promising approach.

The rest of this paper is structured as follows.
In Section~\ref{sec:MF}, we formulate the RL problem and review model-free RL methods
including PGPE.
We then propose the model-based PGPE method in Section~\ref{sec:MB},
and experimentally demonstrate its usefulness in Section~\ref{sec:experiments}.
Finally, we conclude in Section~\ref{sec:conclusion}.

\section{Problem Formulation and Model-Free Policy Search}
\label{sec:MF}
In this section, we first formulate our RL problem
and review existing model-free policy search methods.

\subsection{Formulation}
\label{subsec:formulation}

Let us consider a Markov decision problem consisting of the following elements:
\begin{itemize}
\item $\S$: A set of continuous states.
\item $\A$: A set of continuous actions.
\item $\pIs{\s}$: The (unknown) probability density of initial states.
\item $\psa{\s'}{\s}{\a}$: The (unknown) conditional probability density of
visiting state $\s'$ from state $\s$ by action $\a$.
\item $\Rsa{\s}{\a}{\s'}$: The immediate reward function
for the transition from $\s$ to $\s'$ by $\a$.
\end{itemize}

Let $\pa{\a}{\s}{\para}$ be a policy of an agent parameterized by $\para$,
which is the conditional probability density of taking action $\a$ at state $\s$.
Let
\begin{align*}
  \h:=[\s_1,\a_1,\ldots,\s_\Nstep,\a_\Nstep,\st{\Nstep+1}]
\end{align*}
be a history, which is a sequence of states and actions with finite length $\Nstep$
generated as follows:
First, the initial state $\s_1$ is determined following
the initial-state probability density $\pIs{\s}$.
Then action $\a_1$ is chosen following policy $\pa{\a}{\s}{\para}$,
and next state $\s_2$ is determined following the transition probability density
$\psa{\s'}{\s}{\a}$.
This process is repeated $\Nstep$ times.

Let $\rh{\h}$ be the return for history $\h$,
which is the discounted sum of future rewards the agent can obtain:
\begin{align*}
\rh{\h}:=\sum^{\Nstep}_{\nstep=1}\rdis^{\nstep-1}\Rsa{\st{\nstep}}{\at{\nstep}}{\st{\nstep+1}},
\end{align*}
where $\rdis\in(0,1]$ is a discount factor.
The expected return is given by
\begin{align*}
\J{\para}:=\int\rh{\h}\ph{\h}{\para}\mathrm{d}\h,
\end{align*}
where $\ph{\h}{\para}$ is the probability density of observing history $\h$:
\begin{align*}
\ph{\h}{\para}=\pIs{\st{1}}\prod_{\nstep=1}^{\Nstep}
\pa{\at{\nstep}}{\st{\nstep}}{\para}\psa{\st{\nstep+1}}{\st{\nstep}}{\at{\nstep}}.
\end{align*}

The goal of RL is to find optimal policy parameter $\optpara$ that
maximizes the expected return $\J{\para}$:
\begin{align*}
\optpara:=\argmax_{\para}\J{\para}.
\end{align*}

\subsection{REINFORCE}
\label{subsec:reinforce}

REINFORCE \cite{mach:Williams:1992} is a classic method
for learning the policy parameter $\para$ via gradient ascent:
\begin{align*}
\para\leftarrow\para+\lrate\gradp\J{\para},
\end{align*}
where $\lrate>0$ denotes the learning rate
and $\gradp\J{\para}$ denotes the gradient of $\J{\para}$ with respect to $\para$.

The gradient $\gradp\J{\para}$ can be expressed as
\begin{align*}
\gradp\J{\para}&=\int\rh{\h}\gradp\ph{\h}{\para}\mathrm{d}\h\\
&=\int\rh{h}\ph{\h}{\para}\gradp\log\ph{\h}{\para}\mathrm{d}\h\\
&=\int\rh{h}\ph{\h}{\para}\sum_{\nstep=1}^{\Nstep}
\gradp\log\pa{\at{\nstep}}{\st{\nstep}}{\para}\mathrm{d}{\h},
\end{align*}
where we used
\begin{align*}
\gradp\ph{\h}{\para}=\ph{\h}{\para}\gradp\log\ph{\h}{\para}.
\end{align*}

In the above expression, the probability density of histories, $\ph{\h}{\para}$, is unknown.
Suppose that we are given $\Nepi$ roll-out samples $\{\hn{\nepi}\}_{\nepi=1}^{\Nepi}$
for the current policy, where
\begin{align*}
  \hn{\nepi}=[\snt{\nepi}{1},\ant{\nepi}{1},\ldots,\snt{\nepi}{\Nstep},
  \ant{\nepi}{\Nstep},\snt{\nepi}{\Nstep+1}].
\end{align*}
Then the expectation over $\ph{\h}{\para}$
can be approximated by the empirical average over the samples $\{\hn{\nepi}\}_{\nepi=1}^{\Nepi}$,
i.e., an empirical approximation of the gradient $\gradp\J{\para}$ is given by
\begin{align*}
  \gradp\Jhat{\para}:=\frac{1}{\Nepi}\sum_{\nepi=1}^{\Nepi}\rh{\hn{\nepi}}
  \sum_{\nstep=1}^{\Nstep}\gradp\log\pa{\ant{\nepi}{\nstep}}{\snt{\nepi}{\nstep}}{\para}.
\end{align*}
It is known \cite{IROS:Peters+Schaal:2006}
that the variance of the above gradient estimator can be reduced
by subtracting the baseline $\bline$:
\begin{align*}
  \gradp\Jhatb{\para}{\bline}:=\frac{1}{\Nepi}\sum_{\nepi=1}^{\Nepi}(\rh{\hn{\nepi}}-\bline)
  \sum_{\nstep=1}^{\Nstep}\gradp\log\pa{\ant{\nepi}{\nstep}}{\snt{\nepi}{\nstep}}{\para},
\end{align*}
where
\begin{align*}
  \bline
  &=\frac{\frac{1}{\Nepi}\sum_{\nepi=1}^{\Nepi}{\rh{\hn{\nepi}}
      \unorms{\grad{\para}\log\pdist{\hn{\nepi}}{\para}}}}
  {\frac{1}{\Nepi}\sum_{\nepi=1}^{\Nepi}{\unorms{\grad{\para}\log\pdist{\hn{\nepi}}{\para}}}}.
\end{align*}

Let us consider the following Gaussian policy model with
policy parameter $\para=(\pmean,\pstd^2)$:
\begin{align*}
  \pa{\a}{\s}{\para}=\frac{1}{\sqrt{2\pi\pstd^2}}
    \exp\left(-\frac{(\a-\pmean^\top\cbasis(\s))^2}{2\pstd^2}\right),
\end{align*}
where $^\top$ denotes the transpose, $\pmean$ is the Gaussian mean,
$\pstd$ is the Gaussian standard deviation,
and $\cbasis(\s)$ is the basis function vector.
Then the policy gradients are explicitly expressed as
\begin{align*}
  \grad{\pmean}\log\pa{\a}{\s}{\para}&=\frac{\a-\pmean^\top\cbasis(\s)}{\pstd^2}\cbasis(\s),\\
  \grad{\pstd}\log\pa{\a}{\s}{\para}&=\frac{(a-\pmean^\top\cbasis(\s))^2-\pstd^2}{\pstd^3}.
\end{align*}

REINFORCE is a simple policy-search algorithm
that directly updates policies to increase the expected return.
However, gradient estimates tend to have large variance even if
it is combined with variance reduction by baseline subtraction.
For this reason, policy update by REINFORCE tends to be unreliable
\cite{IROS:Peters+Schaal:2006}.
In particular, the variance of gradient estimates in REINFORCE
can be proportional to the length of the history, $\Nstep$,
due to the stochasticity of policies \cite{NN:Zhao+etal:2012}.
This can be a critical limitation
when the history is long.

\subsection{Policy Gradients with Parameter-Based Exploration (PGPE)}
\label{subsec:PGPE}

To overcome the above limitation of REINFORCE,
a novel policy-search method called \emph{policy gradients with parameter-based exploration} (PGPE)
was proposed recently \cite{NN:Sehnke+etal:2010}.
In PGPE, a deterministic policy (such as the linear policy) is adopted,
and the stochasticity for exploration
is introduced by drawing the policy parameter $\para$ from
a prior distribution $\pdist{\para}{\hpara}$ with hyper-parameter $\hpara$.
Thanks to this per-trajectory formulation, the variance of gradient estimates
can be drastically reduced.

In the PGPE formulation,
the expected return is represented as a function of the hyper-parameter $\hpara$:
\begin{align*}
  \J{\hpara}&:=\iint\rh{\h}\ph{\h}{\para}\pdist{\para}{\hpara}\mathrm{d}\h \mathrm{d}\para.
\end{align*}
Differentiating this with respect to $\hpara$, we have
\begin{align*}
  \grad{\hpara}\J{\hpara}&=\iint\rh{\h}\ph{\h}{\para}\grad{\hpara}\pdist{\para}{\hpara}
  \mathrm{d}\h \mathrm{d}\para\\
  &=\iint\rh{h}\ph{\h}{\para}\pdist{\para}{\hpara}\grad{\hpara}\log\ph{\para}{\hpara}
  \mathrm{d}\h \mathrm{d}\para.
\end{align*}

Because of the per-trajectory formulation, roll-out samples
in the PGPE framework are accompanied with policy parameters,
i.e., $\{(\hn{\nepi},\paran{\nepi})\}_{\nepi=1}^{\Nepi}$.
Based on these paired samples,
an empirical estimator of the above gradient (with baseline subtraction)
is given as follows \cite{NN:Zhao+etal:2012}:
\begin{align*}
  \grad{\hpara}\Jhatb{\hpara}{\bline}:=\frac{1}{\Nepi}\sum_{\nepi=1}^{\Nepi}
  (\rh{\hn{\nepi}}-\bline)\grad{\hpara}\log\pdist{\paran{\nepi}}{\hpara},
\end{align*}
where
\begin{align*}
  \bline
  &=\frac{\frac{1}{\Nepi}\sum_{\nepi=1}^{\Nepi}{\rh{\hn{\nepi}}
      \unorms{\grad{\hpara}\log\pdist{\paran{\nepi}}{\hpara}}}}
  {\frac{1}{\Nepi}\sum_{\nepi=1}^{\Nepi}{
      \unorms{\grad{\hpara}\log\pdist{\paran{\nepi}}{\hpara}}}}.
\end{align*}

Let us employ
the linear deterministic policy, i.e., action $\a$ is chosen as $\para^\top\cbasis(\s)$
for some basis function $\cbasis$.
The parameter vector $\para$ is drawn from the Gaussian prior distribution
with hyper-parameter $\hpara=(\hmean,\hvar)$.
Here $\hmean$ denotes the Gaussian mean vector and $\hvar$ denotes the vector
consisting of the Gaussian standard deviation in each element:
\begin{align*}
  \pdist{\theta_i}{\rho_i}=\frac{1}{\sqrt{2\pi\tau_i^2}}
  \exp\left(-\frac{(\theta_i-\eta_i)^2}{2\tau_i^2}\right),
\end{align*}
where $\theta_i$, $\rho_i$, $\eta_i$, and $\tau_i$ are
the $i$-th elements of $\para$, $\hpara$, $\hmean$, and $\hvar$, respectively.
Then the derivatives of $\log \pdist{\para}{\hpara}$ with respect to
$\eta_i$ and $\tau_i$ are given as follows:
\begin{align*}
  \grad{\eta_i}\log \pdist{\para}{\hpara}&=\frac{\theta_i-\eta_i}{\tau_i^2},\\
  \grad{\tau_i} \log \pdist{\para}{\hpara}&=\frac{(\theta_i-\eta_i)^2-\tau_i^2}{\tau_i^3}.
\end{align*}

\subsection{Importance-Weighted PGPE (IW-PGPE)}
\label{subsec:IWPGPE}
A popular idea to further improve the performance of RL methods
is to reuse previously collected samples
\cite{book:Sutton+Barto:1998,NN:Hachiya+etal:2009}.
Such a sample-reuse strategy is particularly useful
when data sampling costs is high (e.g., robot control).

\emph{Importance-weighted} PGPE (IW-PGPE) \cite{NC:Zhao+etal:2013}
combines the sample-reuse idea with PGPE.
Technically, IW-PGPE can be regarded as an off-policy extension of PGPE,
where data collecting policies are different from the current policy.
In the PGPE formulation, such a off-policy scenario can be regarded as the situation where
data collecting policies and the current policy are drawn from different
prior distributions (more specifically, different hyper-parameters).
Let $\hpara$ be the hyper-parameter for the current policy
and $\hpara'$ be the hyper-parameter for a data collecting policy.
Let us denote data samples collected with hyper-parameter $\hpara'$ as
$\{(\paran{\nepi}',\hn{\nepi}')\}_{\nepi=1}^{\Nepi'}$.

When the data collecting policy is different from the current policy,
\emph{importance sampling} is a useful technique to correct the estimation bias
caused by differing distributions \cite{book:Sugiyama+Kawanabe:2012}.
More specifically, the gradient is estimated as
\begin{align*}
\grad{\hpara}\Jhatbw{\bline}{w}{\hpara}{}=
\frac{1}{\Nepi'}\sum_{\nepi=1}^{\Nepi'}w(\paran{\nepi}')(\rh{\hn{\nepi}'}-\bline)
\grad{\hpara}\log\pdist{\paran{\nepi}'}{\hpara},
\end{align*}
where $w(\para)$ is the \emph{importance weight} defined as
\begin{align*}
w(\para):=\frac{\pdist{\para}{\hpara}}{\pdist{\para}{\hpara'}},
\end{align*}
and $\bline$ is the baseline given by
\begin{align*}
\bline
&=\frac{\frac{1}{\Nepi'}\sum_{\nepi=1}^{\Nepi'}
\rh{\hn{\nepi}'}w^2(\paran{\nepi}')\unorms{\grad{\hpara}\log\pdist{\paran{\nepi}'}{\hpara}}}
{\frac{1}{\Nepi'}\sum_{\nepi=1}^{\Nepi'}
w^2(\paran{\nepi}')\unorms{\grad{\hpara}\log\pdist{\paran{\nepi}'}{\hpara}}}.
\end{align*}
Through experiments,
the IW-PGPE method was demonstrated to be the best performing algorithm
in model-free RL approaches \cite{NC:Zhao+etal:2013}.

The purpose of this paper is to develop a model-based counterpart of PGPE.

\section{Model-Based Policy Search}
\label{sec:MB}
Model-based RL first estimates the transition model
and then learns a policy based on the estimated transition model.
Because one can draw as many trajectory samples as one wants
from the learned transition model without additional sampling costs,
the model-based approach can work well if the transition model is accurately estimated
 \cite{Wang:2003,Deisenroth:2011}.
In this section, we extend PGPE to a model-based scenario.
We first review an existing model estimation method
based on the Gaussian process (GP) \cite{book:Rasmussen+Williams:2006}
and point out its limitations.
Then we propose to use the state-of-the-art conditional density estimator
called \emph{least-squares conditional density estimation} (LSCDE) \cite{IEICE:Sugiyama+etal:2010a}
in the model-based PGPE method.

\subsection{Model-Based PGPE (M-PGPE)}
\label{subsec:prop}

PGPE can be extended to a model-based scenario as follows.

\begin{enumerate}
\item Collect transition samples $\{(\s_m,\a_m,\s'_m)\}_{m=1}^M$.
\item Obtain transition model $\phat{\s'|\s,\a}$ by a model estimation method
  from $\{(\s_m,\a_m,\s'_m)\}_{m=1}^M$.
\item Initialize hyper-parameter $\hpara$.
\item Draw policy parameter $\para$ from prior distribution $\pdist{\para}{\hpara}$.
\item Generate many samples $\{\widetilde{h}_{n}\}_{n=1}^{\widetilde{N}}$
  from $\phat{\s'|\s,\a}$ and current policy $\pa{\a}{\s}{\para}$.
\item Estimate baseline $\bline$ and gradient $\grad{\hpara}\Jhatb{\hpara}{\bline}$
  from disjoint subsets of $\{\widetilde{h}_{n}\}_{n=1}^{\widetilde{N}}$.
\item Update hyper-parameter as
  $\hpara\leftarrow\hpara+\lrate\grad{\hpara}\Jhatb{\hpara}{\bline}$,
  where $\lrate>0$ denotes the learning rate.
\item Repeat Steps 4--7 until $\hpara$ converges.
\end{enumerate}


Below, we consider the problem of approximating
the transition probability $\psa{\s'}{\s}{\a}$
from samples $\{(\s_m,\a_m,\s'_m)\}_{m=1}^M$,
and review transition model estimation methods.

\subsection{Gaussian Process (GP)}
Here we review a transition model estimation method based on GP.


In the GP framework, the problem of transition probability estimation is formulated
as the regression problem of predicting output $\s'$ given input $\s$ and $\a$
under Gaussian noise:
\[
	\s' = f(\s, \a) + \varepsilon,
\]
where $f$ is an unknown regression function and $\varepsilon\sim \mathcal{N}(0,\sigma_{\varepsilon}^2)$ is independent Gaussian noise.
Then, the GP estimate of the transition probability density $\psa{\s'}{\s}{\a}$
for an arbitrary test input $\s$ and $\a$
is given by the Gaussian distribution with mean and variance given by
\begin{align*}
\boldsymbol{k}^\top(\boldsymbol{K}+\sigma_{\varepsilon}^2 \boldsymbol{I}_M)^{-1}\boldsymbol{y}
~~~\mbox{and}~~~
k(\s,\a,\s,\a)-\boldsymbol{k}^\top(\boldsymbol{K}+\sigma_{\varepsilon}^2 \boldsymbol{I}_M)^{-1}\boldsymbol{k},
\end{align*}
respectively.
Here, $\I{M}$ denotes the $M$-dimensional identity matrix.
$\boldsymbol{k}$ is the $M$-dimensional vector
and $\boldsymbol{K}$ is the $M\times M$ Gram matrix defined by
\begin{align*}
  \boldsymbol{k}&=
  \begin{pmatrix}
    k(\s_1,\a_1,\s,\a)\\
    \vdots\\
    k(\s_M,\a_M,\s,\a)
  \end{pmatrix}
  ~~~\mbox{and}~~~
  \boldsymbol{K}=
  \begin{pmatrix}
 k(\s_1,\a_1,\s_1,\a_1)&\ldots&k(\s_M,\a_M,\s_1,\a_1)\\
 \vdots&\ddots&\vdots\\
 k(\s_1,\a_1,\s_M,\a_M)&\ldots&k(\s_M,\a_M,\s_M,\a_M)\\
  \end{pmatrix}.
\end{align*}
$k(\s,\a,\s',\a')$ denotes the covariance function, which is, e.g., defined by
\[
k(\s,\a,\s',\a') = \theta \exp\left(-([\s^\top,\a]-[\s'^\top,\a'])\boldsymbol{\Theta}([\s^\top,\a]-[\s'^\top,\a'])^\top\right).
\]
Here, $\theta$ and $\boldsymbol{\Theta}$ are hyperparameters,
and together with the noise variance $\sigma_{\varepsilon}^2$,
the hyperparameters
are determined by evidence maximization \cite{book:Rasmussen+Williams:2006}.

As shown above, the GP-based model estimation method
requires the strong assumption that
the transition probability density $\psa{\s'}{\s}{\a}$ is Gaussian.
That is, GP is non-parametric as a regression method of estimating
the conditional mean, it is parametric (Gaussian) as a conditional density estimator.
Such a conditional Gaussian assumption is highly restrictive in RL problems.

\subsection{Least-Squares Conditional Density Estimation (LSCDE)}
\label{subsec:LSCDE}
To overcome the restriction of the GP-based model estimation method,
we propose to use LSCDE.

Let us model the transition probability $\psa{\s'}{\s}{\a}$
by the following linear-in-parameter model:
\begin{align}
\qsa{\s'}{\s}{\a}&:=\trans{\cparam}\cbasis(\s,\a,\s')\nonumber\\
&\phantom{:}=
\sum_{m=1}^M
\alpha_m\exp\left(-\frac{\|\s-\s_m\|^2}{2\kappa^2}\right)
\exp\left(-\frac{(\a-\a_m)^2}{2\kappa^2}\right)
\exp\left(-\frac{\|\s'-\s'_m\|^2}{2\kappa^2}\right),
\label{LSCDE-model}
\end{align}
where $\cbasis(\s,\a,\s')$ is the $M$-dimensional basis function vector
and $\cparam$ is the $M$-dimensional parameter vector.
If $M$ is too large, we may reduce the number of basis functions
by only using a subset of samples as Gaussian centers.
We may use different Gaussian widths for $\s$ and $\a$ if necessary.

The parameter $\cparam$ in the model \eqref{LSCDE-model}
is learned so that the following squared error is minimized:
\begin{align*}
\Gz{\cparam}:=\frac{1}{2}\iiint\Big(\qsa{\s'}{\s}{\a}-\psa{\s'}{\s}{\a}\Big)^2
p(\s,\a)\mathrm{d}\s\mathrm{d}\a\mathrm{d}\s'.
\end{align*}
This can be expressed as
\begin{align*}
\Gz{\cparam}
&=\frac{1}{2}\iiint\qsa{\s'}{\s}{\a}^2p(\s,\a)\mathrm{d}\s\mathrm{d}\a\mathrm{d}\s'\\
&\phantom{=}
-\iiint\qsa{\s'}{\s}{\a}p(\s,\a,\s')\mathrm{d}\s\mathrm{d}\a\mathrm{d}\s'+C\\
&=\frac{1}{2}\iiint\Big(\trans{\cparam}\cbasis(\s,\a,\s')\Big)^2
p(\s,\a)\mathrm{d}\s\mathrm{d}\a\mathrm{d}\s'\\
&\phantom{=}
-\iiint\trans{\cparam}\cbasis(\s,\a,\s')p(\s,\a,\s')\mathrm{d}\s\mathrm{d}\a\mathrm{d}\s'+C,
\end{align*}
where we used $\psa{\s'}{\s}{\a}=p(\s,\a,\s')/p(\s,\a)$ in the second term
and
\begin{align*}
C:=\frac{1}{2}\iiint\psa{\s'}{\s}{\a}p(\s,\a,\s')\mathrm{d}\s\mathrm{d}\a\mathrm{d}\s'.
\end{align*}
Because $C$ is constant, we only consider the first two terms from here on:
\begin{align*}
\G{\cparam}&:=\Gz{\cparam}-C\\
&\phantom{:}=\frac{1}{2}\trans{\cparam}\cbasismat\cparam-\trans{\cparam}\cbasisvec,
\end{align*}
where
\begin{align*}
\cbasismat&:=\iint\cbasisbar(\s,\a)p(\s,\a)\mathrm{d}\s\mathrm{d}\a
\in\mathbb{R}^{M\times M},\\
\cbasisvec&:=\iiint\cbasis(\s,\a,\s')p(\s,\a,\s')\mathrm{d}\s\mathrm{d}\a\mathrm{d}\s'
\in\mathbb{R}^{M},\\
\cbasisbar(\s,\a)&:=\int\cbasis(\s,\a,\s')\trans{\cbasis(\s,\a,\s')}\mathrm{d}\s'
\in\mathbb{R}^{M\times M}.
\end{align*}
Note that, for the Gaussian model \eqref{LSCDE-model},
the $(m,m')$-th element of $\cbasisbar(\s,\a)$ can be computed analytically as
\begin{align*}
\cbasisbar_{m,m'}(\s,\a)
&=(\sqrt{\pi}\kappa)^{\mathrm{dim}(\s')}
\exp\left(-\frac{\|\s-\s_m\|^2+\|\s-\s_{m'}\|^2}{2\kappa^2}\right)\\
&\phantom{=}\times
\exp\left(-\frac{(\a-\a_m)^2+(\a-\a_{m'})^2}{2\kappa^2}\right)
\exp\left(-\frac{\|\s'_m-\s'_{m'}\|^2}{4\kappa^2}\right).
\end{align*}

Because $\cbasismat$ and $\cbasisvec$ included in $\G{\cparam}$ contain
the expectations over unknown densities $p(\s,\a)$ and $p(\s,\a,\s')$,
they are approximated by sample averages.
Then we have
\begin{align*}
\Ghat{\cparam}:=\frac{1}{2}\trans{\cparam}\cbasismathat\cparam-\trans{\cbasisvechat}\cparam,
\end{align*}
where
\begin{align*}
\cbasismathat&:=\frac{1}{M}\sum^{M}_{m=1}\cbasisbar(\s_m,\a_m)\in\mathbb{R}^{M\times M},\\
\cbasisvechat&:=\frac{1}{M}\sum^{M}_{m=1}\cbasis(\s_m,\a_m,\s'_m)\in\mathbb{R}^{M}.
\end{align*}

By adding an $\ell_2$-regularizer to $\Ghat{\cparam}$
to avoid overfitting, the LSCDE optimization criterion is given as
\begin{align*}
\cparamtilde:=\argmin_{\cparam\in\mathbb{R}^M}\left[\Ghat{\cparam}+\frac{\lambda}{2}\|\cparam\|^2\right],
\end{align*}
where $\lambda$ $(\ge0)$ is the regularization parameter.
Taking the derivative of the above objective function and equating it to zero,
we can see that the solution $\cparamtilde$ can be obtained
just by solving the following system of linear equations:
\begin{align*}
(\cbasismathat+\lambda\I{M})\cparam=\cbasisvechat,
\end{align*}
where $\I{M}$ denotes the $M$-dimensional identity matrix.
Thus, the solution $\cparamtilde$ is given analytically as
\begin{align*}
\cparamtilde=(\cbasismathat+\lambda\I{M})^{-1}\cbasisvechat.
\end{align*}
Because conditional probability densities are non-negative by definition,
we modify the solution $\cparamtilde$ as
\begin{align*}
\cparamhat:=\max(\boldsymbol{0}_M,\cparamtilde),
\end{align*}
where $\boldsymbol{0}_M$ denotes the $M$-dimensional zero vector
and `$\max$' for vectors are applied in the element-wise manner.

Finally, we renormalize the solution in the test phase.
More specifically, given a test input point $(\s,\a)$,
the final LSCDE solution is given as
\begin{align}
\phat{\s'|\s,\a}=\frac{\trans{\cparamhat}\cbasis(\s,\a,\s')}
{\int\trans{\cparamhat}\cbasis(\s,\a,\s')\mathrm{d}\s'},
\label{eq:est-p}
\end{align}
where, for the Gaussian model \eqref{LSCDE-model},
the denominator in Eq.\eqref{eq:est-p} can be analytically computed as
\begin{align*}
  \int\trans{\cparamhat}\cbasis(\s,\a,\s')\mathrm{d}\s'
=(\sqrt{2\pi}\kappa)^{\mathrm{dim}(\s')}
\sum_{m=1}^M\alpha_m
\exp\left(-\frac{\|\s-\s_m\|^2+(\a-\a_m)^2}{2\kappa^2}\right).
\end{align*}
LSCDE was proved to achieve the optimal non-parametric convergence rate
to the true conditional density in the mini-max sense \cite{IEICE:Sugiyama+etal:2010a},
meaning that no method can outperform this simple LSCDE method asymptotically.

Model selection of the Gaussian width $\kappa$ and the regularization parameter $\lambda$
is possible by cross-validation.
A MATLAB$^\text{\textregistered}$ implementation of
LSCDE is available from
\begin{center}
`\url{http://sugiyama-www.cs.titech.ac.jp/~sugi/software/LSCDE/}'.
\end{center}

\section{Experiments}
\label{sec:experiments}

In this section, we demonstrate the usefulness of the proposed
method through experiments.

\subsection{Continuous Chain Walk}
\label{subsec:chain}

\begin{figure}[t]
  \centering
\includegraphics[width=0.6\textwidth,clip]{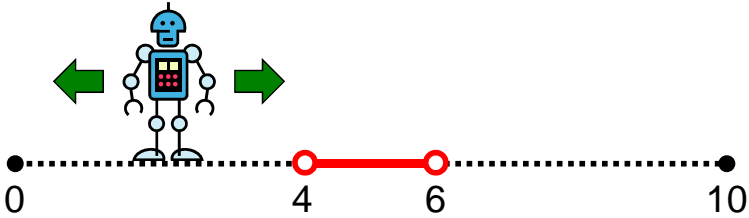}
\caption{Illustration of continuous chain walk.}
\label{fig:cwalk-view}
\end{figure}

For illustration purposes,
let us first consider a simple continuous chain-walk task (Figure~\ref{fig:cwalk-view}).

\subsubsection{Setup}
Let
\begin{align*}
\S&=[0, 10],\\
\A&=[-5, 5],\\
\Rsa{s}{\a}{s'}&=
\begin{cases}
1&(4<s'<6),\\
0&(\mathrm{otherwise}).\\
\end{cases}
\end{align*}
That is, the agent receives positive reward $+1$
at the center of the state space.
We set the episode length at $\Nstep=10$,
the discount factor at $\rdis=0.99$, and the learning rate at $\lrate=0.1$.
We use the following linear-in-parameter policy model:
\begin{align*}
\sum_{i=1}^6\theta_i\exp\left(-\frac{(s-c_i)^2}{2}\right),
\end{align*}
where $(c_1,\ldots,c_6)=(0,2,4,6,8,10)$.

As transition dynamics, we consider two setups:
\begin{description}
\item[Gaussian:] The true transition dynamics is given by
\begin{align*}
s_{t+1}=s_{t}+\a_{t}+\epsilon_t,
\end{align*}
where $\epsilon_t$ is the Gaussian noise with mean $0$ and standard deviation $0.3$.
\item[Bimodal:] The true transition dynamics is given by
\begin{align*}
s_{t+1}=s_{t}\pm\a_{t}+\epsilon_t,
\end{align*}
where $\epsilon_t$ is the Gaussian noise with mean $0$ and standard deviation $0.3$,
and the sign of $a_{t}$ is randomly chosen with probability $1/2$.
\end{description}

We compare the following three policy search methods:
\begin{description}
\item[M-PGPE(LSCDE):] The model-based PGPE method with transition model estimated by LSCDE.
\item[M-PGPE(GP):] The model-based PGPE method with transition model estimated by GP.
\item[IW-PGPE:] The model-free PGPE method with sample reuse by importance weighting\footnote{
We have also tested the plain PGPE method without importance weighting.
However, this did not perform well in our preliminary experiments,
and thus we decided to omit the results.
} \cite{NC:Zhao+etal:2013}.
\end{description}
Below, we consider the situation where the budget for data collection is limited
to $\Nepi=20$ episodic samples.

\subsubsection{LSCDE Vs.~GP}
\label{subsubsec:model}

When the transition model is learned in the M-PGPE methods,
all $\Nepi=20$ samples are gathered randomly in the beginning at once.
More specifically, the initial state $s_1$ and the action $\a_1$
are chosen from the uniform distributions over $\S$ and $\A$, respectively.
Then the next state $s_2$ and the immediate reward $r_1$ are obtained.
Then the action $\a_2$ is chosen from the uniform distribution over $\A$,
and the next state $s_3$ and the immediate reward $r_2$ are obtained.
This process is repeated until we obtain $r_{\Nstep}$.
This gives a trajectory sample, and we repeat this data generation process
$\Nepi$ times to obtain $\Nepi$ trajectory samples.

Figure~\ref{fig:trans_pure} and Figure~\ref{fig:trans_bi} illustrate
the true transition dynamics and its estimates obtained by LSCDE and GP
in the Gaussian and bimodal cases.
Figure~\ref{fig:trans_pure} shows that
both LSCDE and GP can learn the entire profile of the true transition
dynamics well in the Gaussian case.
On the other hand, Figure~\ref{fig:trans_bi} shows that
LSCDE can still successfully capture the entire profile
of the true transition model well even in the bimodal case,
but GP fails to capture the bimodal structure.

Based on the estimated transition models,
we learn policies by the M-PGPE method.
We generate $1000$ artificial samples for policy gradient estimation and
another $1000$ artificial samples for baseline estimation from the learned transition model.
Then policy is updated based on these artificial samples.
We repeat this policy update step $20$ times.
For evaluating the return of a learned policy,
we use $100$ additional test episodic samples which are not used for policy learning.
Figure~\ref{fig:cwalk-results-pure} and Figure~\ref{fig:cwalk-results-bi}
depict the average performance of learned policies over $100$ runs.
As expected, the GP-based method performs very well in the Gaussian case, but
LSCDE still exhibits reasonably good performance.
In the bimodal case, GP performs poorly and LSCDE gives much better policies than GP.
This illustrates the high flexibility of LSCDE.

\subsubsection{Model-Based Vs.~Model-Free}
Next, we compare the performance of M-PGPE with the model-free IW-PGPE method.

\begin{figure}[p]
  \centering
\subfigure[True transition.]{
\includegraphics[width=0.45\textwidth,clip]{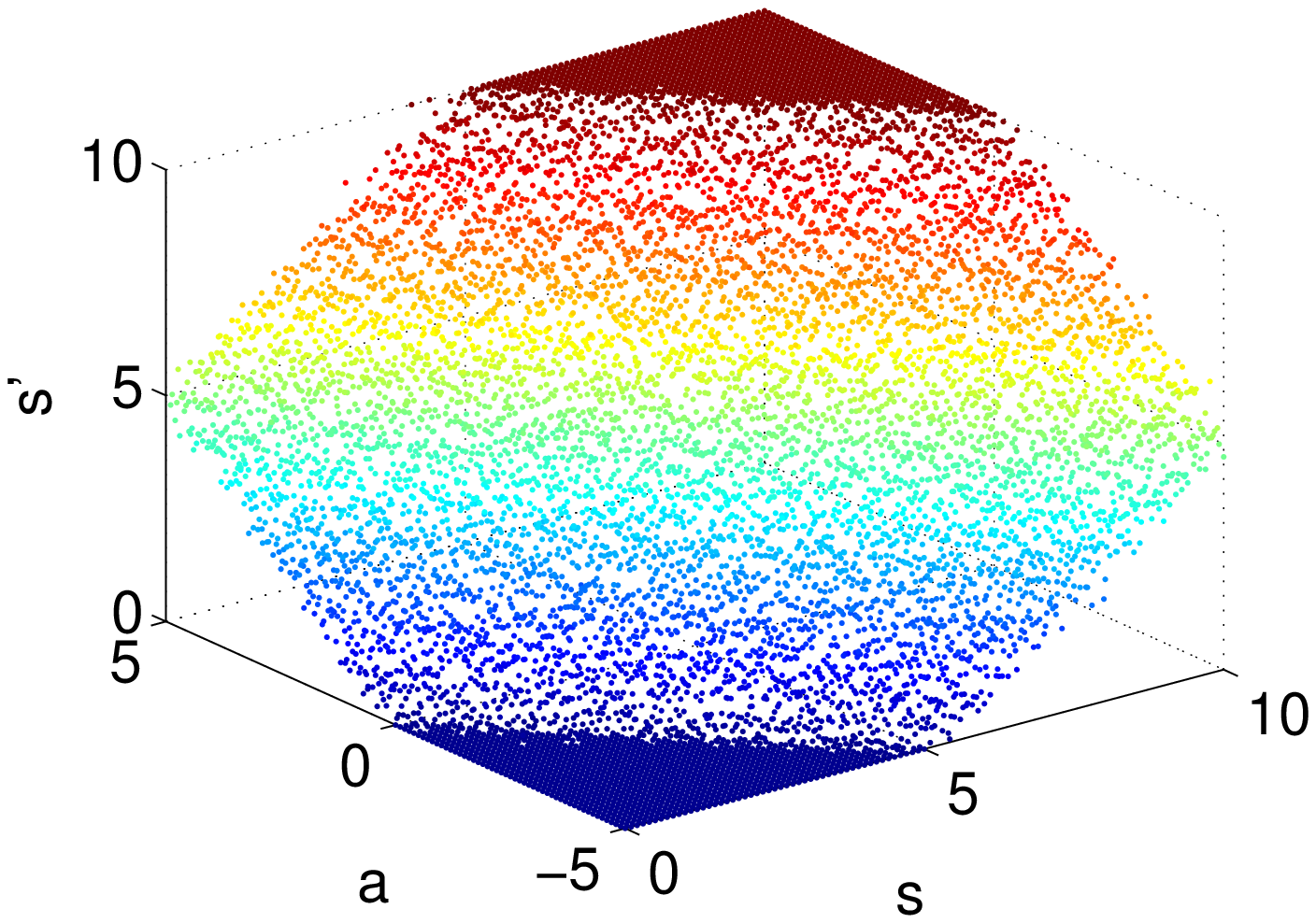}
}\\
\subfigure[Transition estimated by LSCDE.]{
\includegraphics[width=0.45\textwidth,clip]{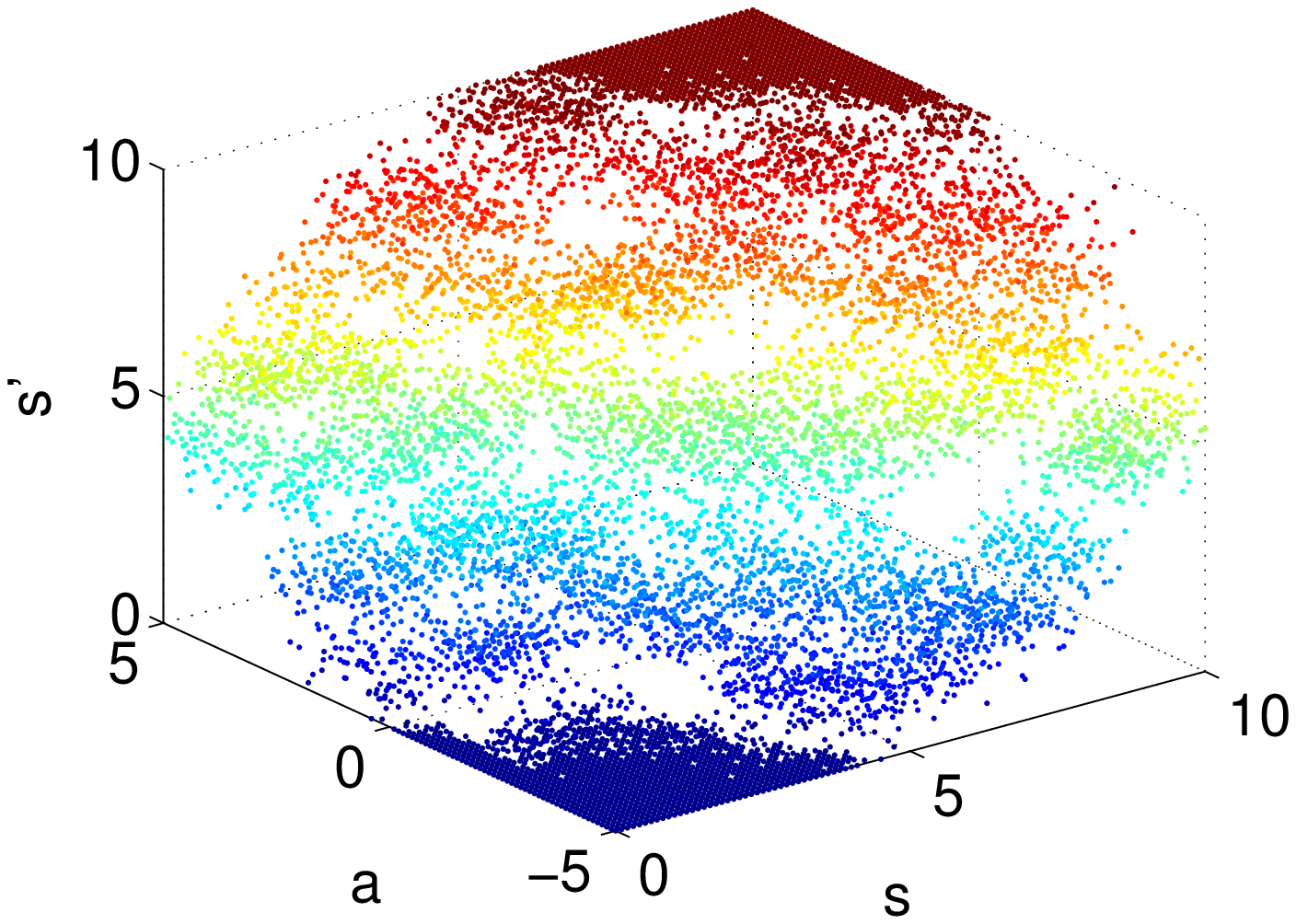}
}
\subfigure[Transition estimated by GP.]{
\includegraphics[width=0.45\textwidth,clip]{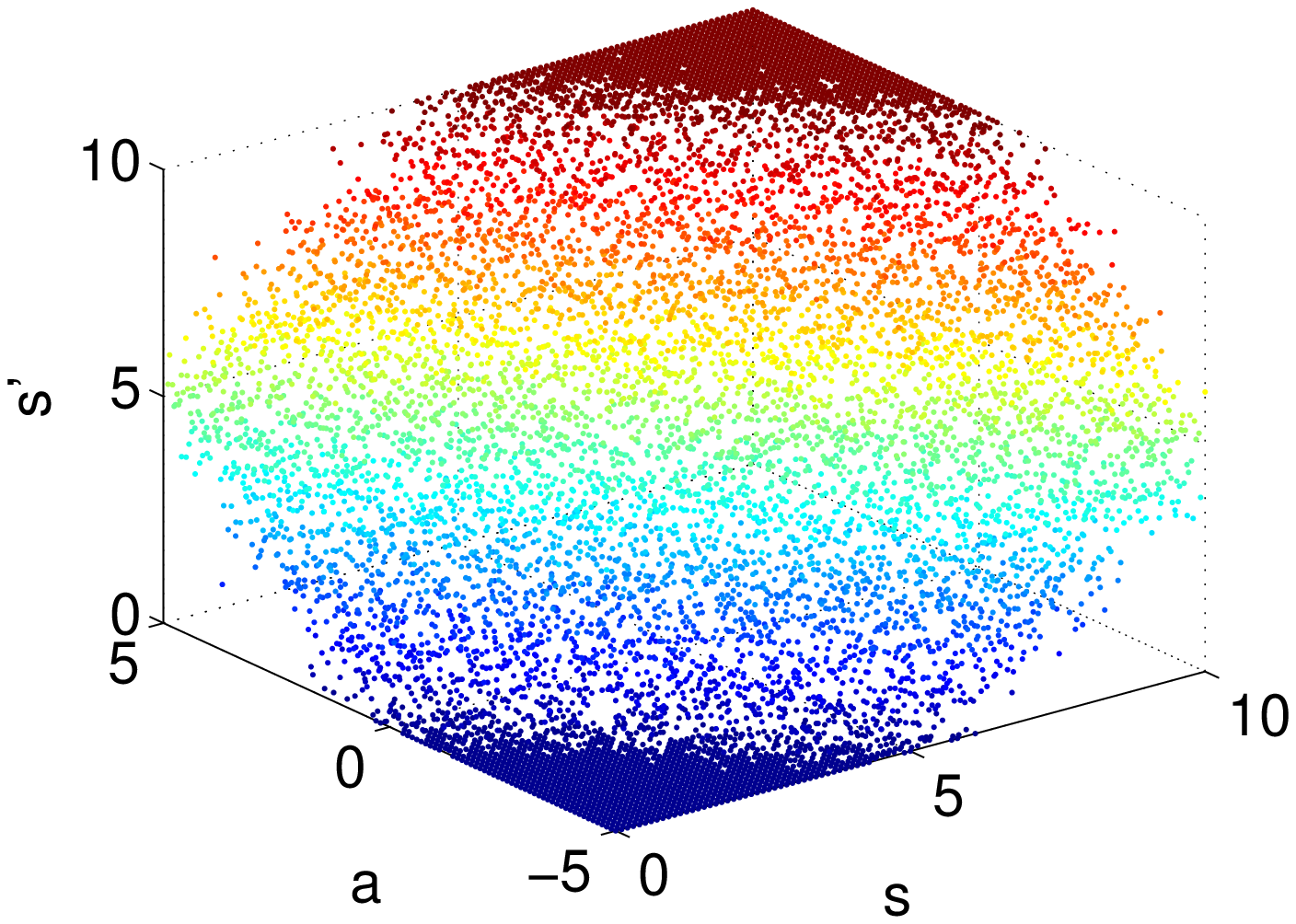}
}
\caption{Gaussian transition dynamics and its estimates by LSCDE and GP.
  $\argmax_{s'}\psa{s'}{s}{\a}$ is plotted as a function of $s$ and $\a$.
}
\label{fig:trans_pure}
\vspace*{5mm}
\begin{minipage}[t]{0.47\textwidth}
\includegraphics[width=1\textwidth,clip]{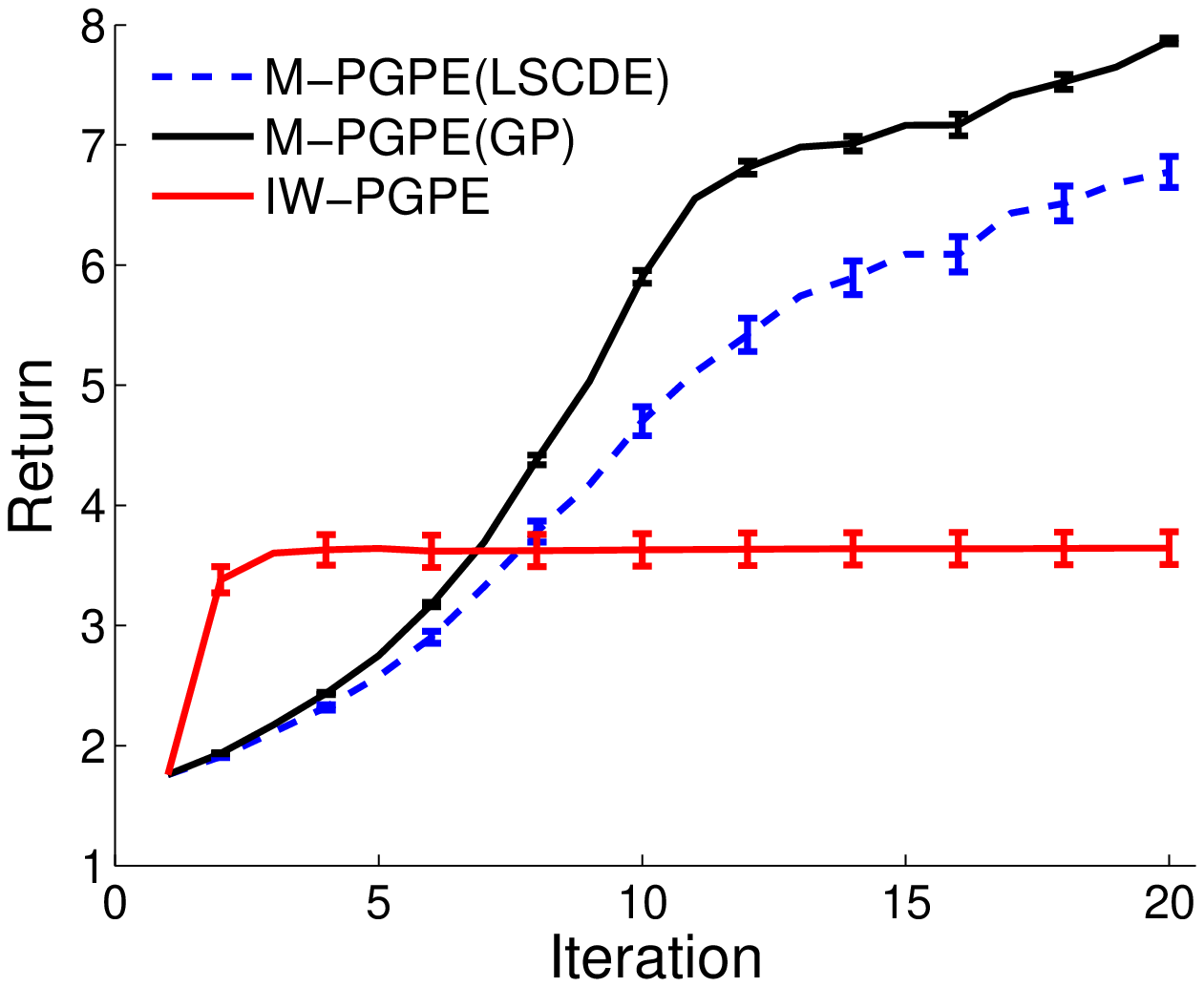}
\caption{Returns of the policies obtained by M-PGPE with LSCDE and GP
as well as IW-PGPE
  for Gaussian transition
(averages and standard errors over $100$ runs).}
\label{fig:cwalk-results-pure}
\end{minipage}
~~
\begin{minipage}[t]{0.47\textwidth}
\includegraphics[width=1\textwidth,clip]{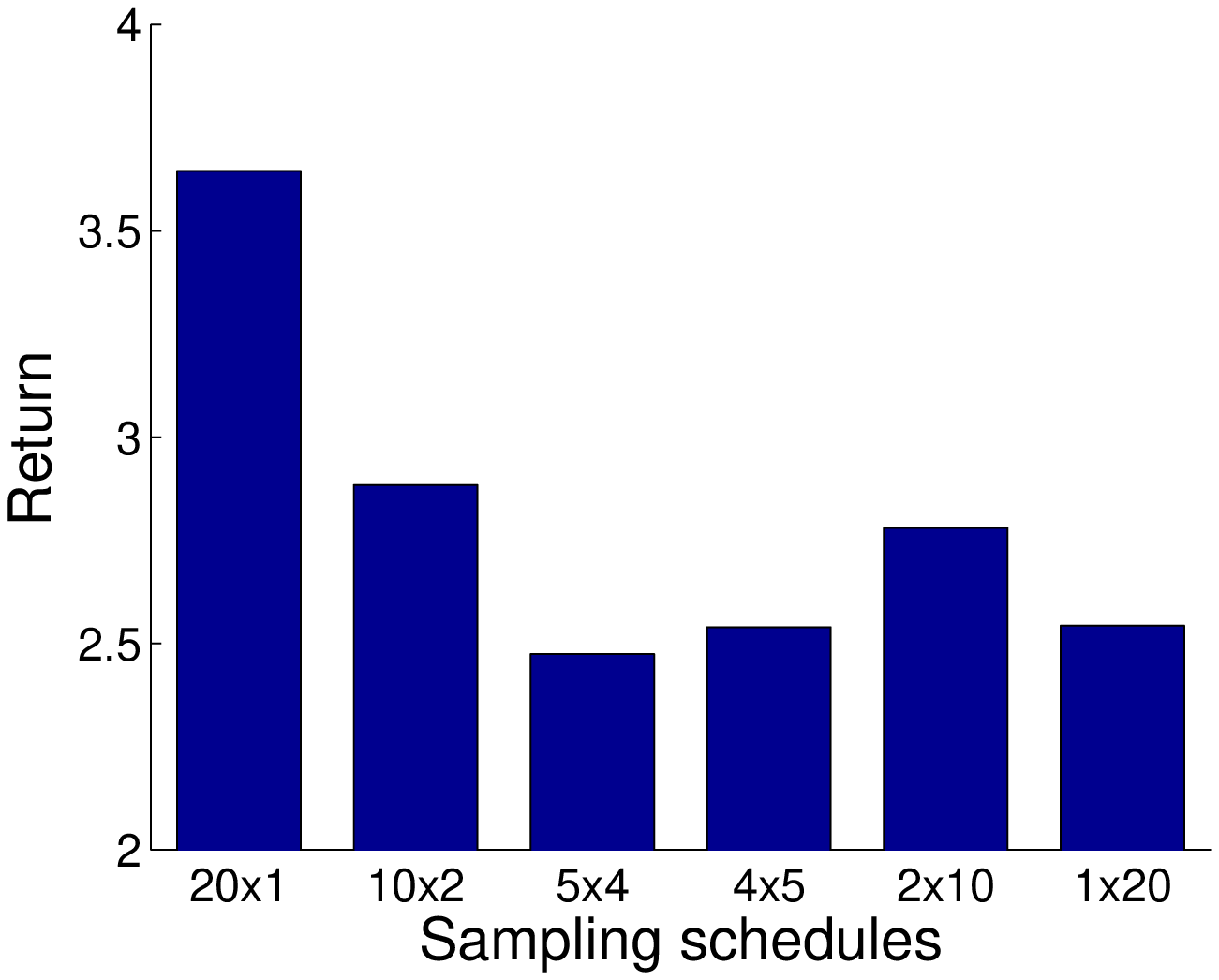}
  \caption{Returns obtained by IW-PGPE averaged over $100$ runs
    for Gaussian transition dynamics with different sampling schedules
    (e.g., $5\times4$ means gathering $k=5$ samples $4$ times).}
  \label{fig:cwalk-schedule-pure}
\end{minipage}
\end{figure}

\begin{figure}[p]
  \centering
\subfigure[True transition.]{
\includegraphics[width=0.45\textwidth,clip]{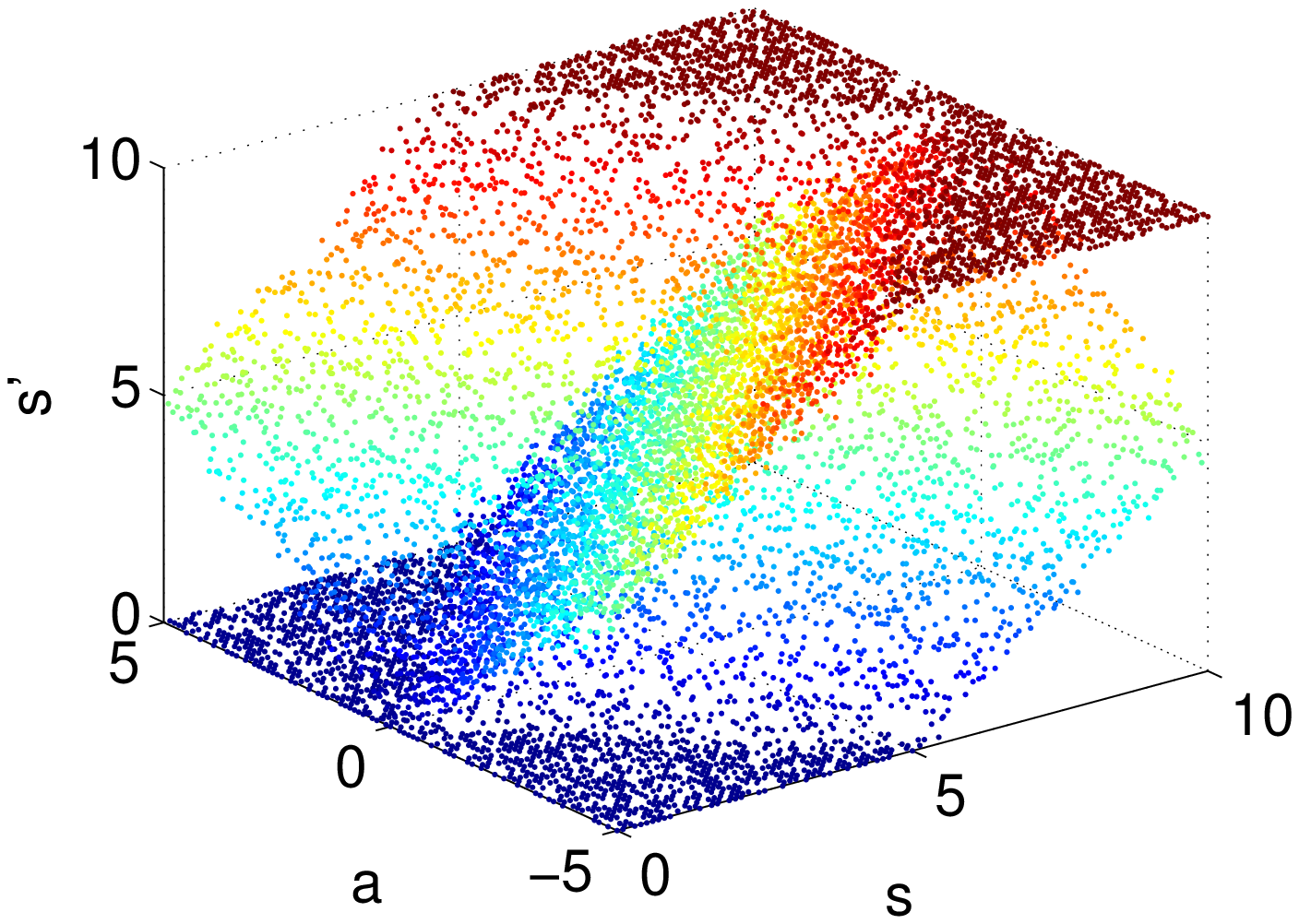}
}\\
\subfigure[Transition estimated by LSCDE.]{
\includegraphics[width=0.45\textwidth,clip]{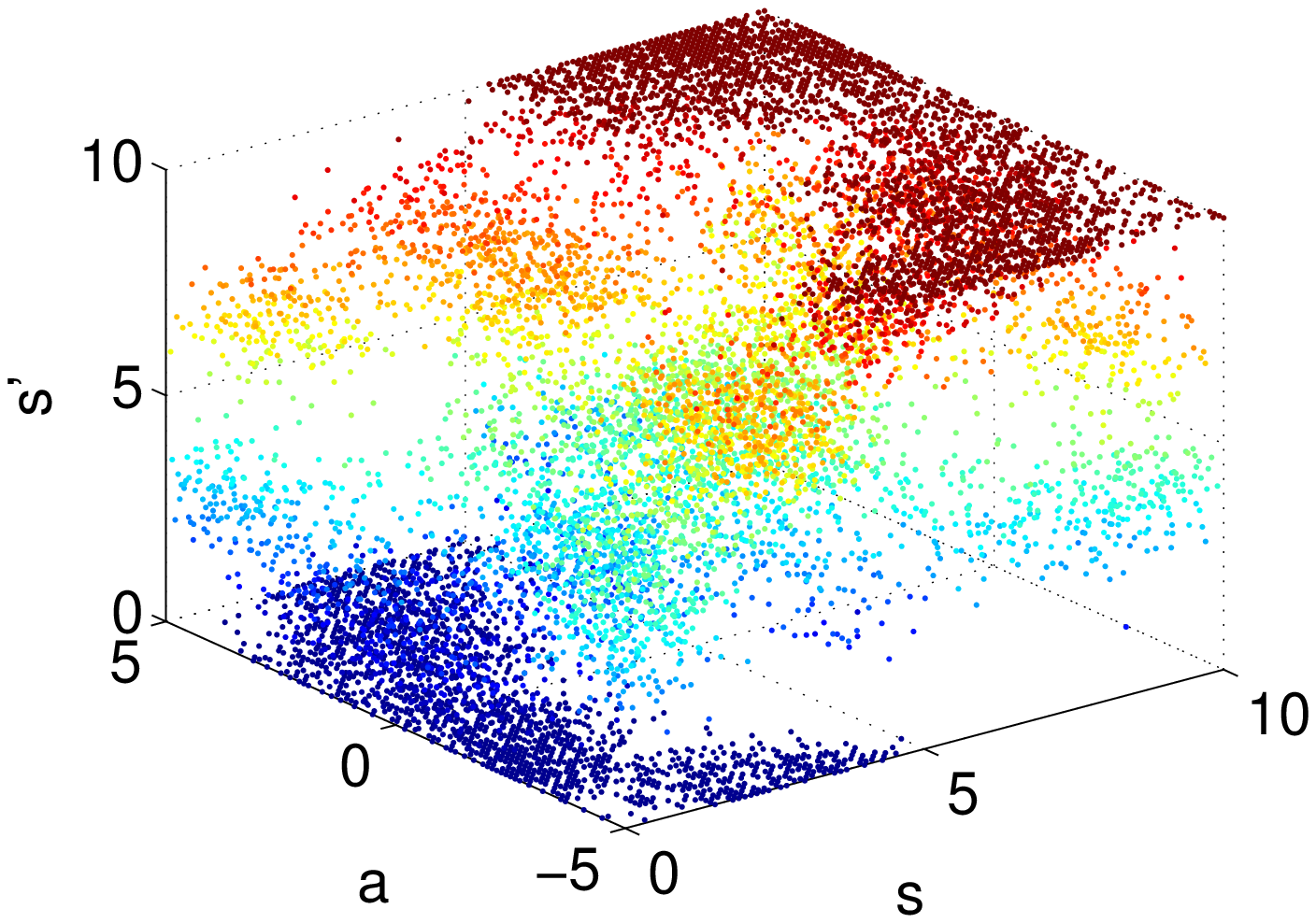}
}
\subfigure[Transition estimated by GP.]{
\includegraphics[width=0.45\textwidth,clip]{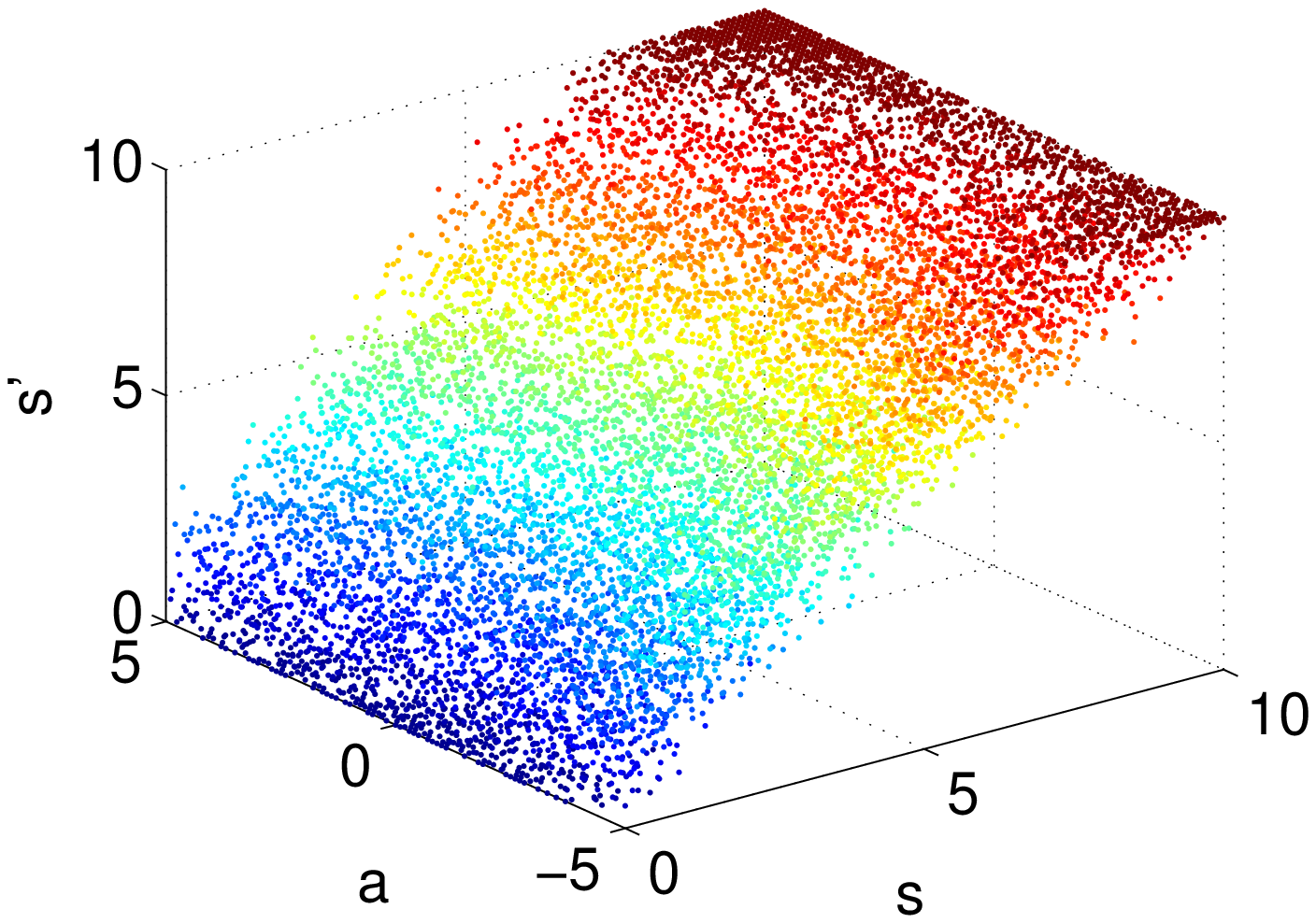}
}
\caption{Bimodal transition dynamics and its estimates by LSCDE and GP.
  $\argmax_{s'}\psa{s'}{s}{\a}$ is plotted as a function of $s$ and $\a$.
}
\label{fig:trans_bi}
\vspace*{5mm}
\begin{minipage}[t]{0.47\textwidth}
\includegraphics[width=\textwidth,clip]{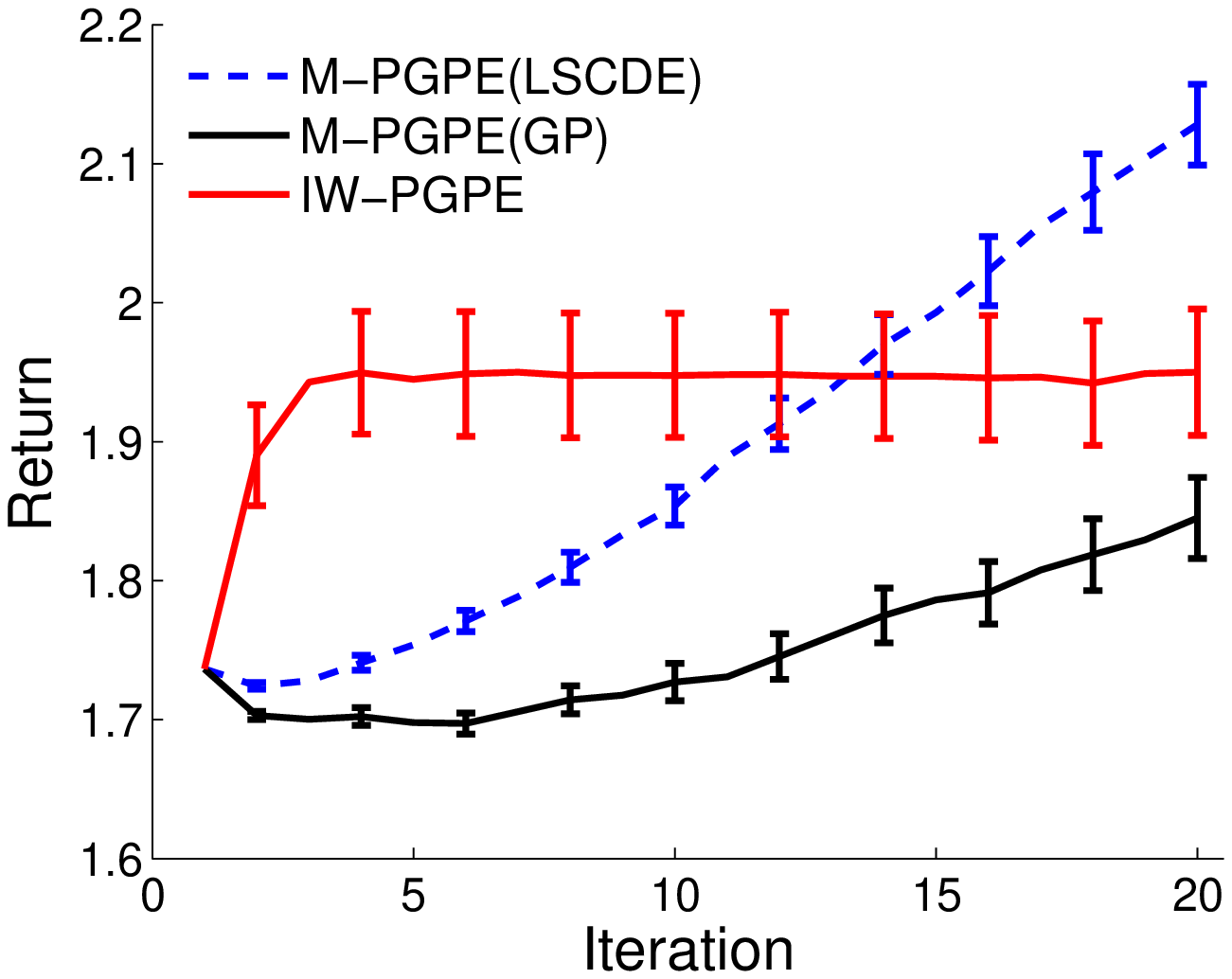}
\caption{Returns of the policies obtained by M-PGPE with LSCDE and GP
as well as IW-PGPE for bimodal transition
(averages and standard errors over $100$ runs).}
\label{fig:cwalk-results-bi}
\end{minipage}
~~
\begin{minipage}[t]{0.47\textwidth}
  \includegraphics[width=\textwidth,clip]{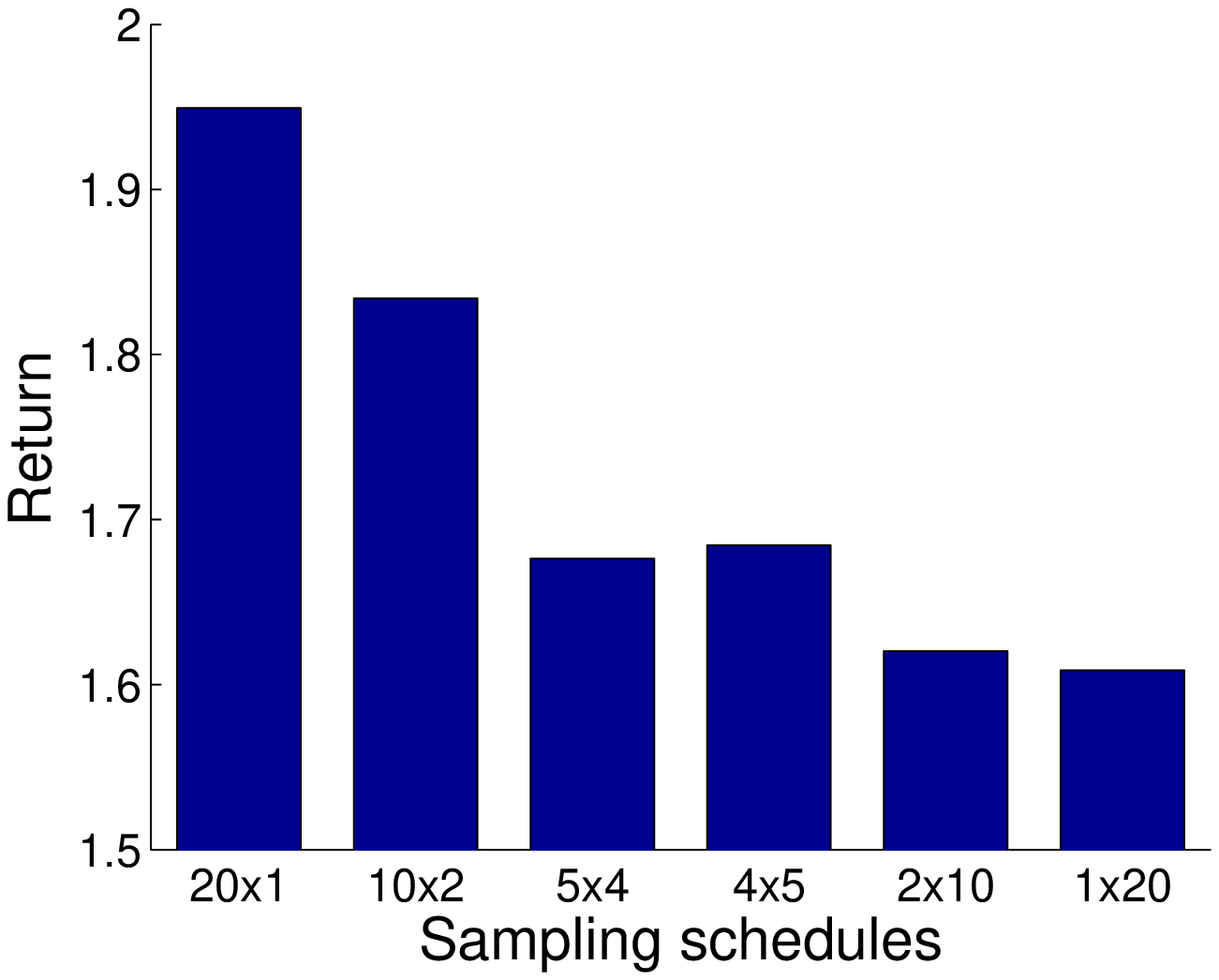}
  \caption{Returns obtained by IW-PGPE averaged over $100$ runs
    for bimodal transition with different sampling schedules
    (e.g., $5\times4$ means gathering $k=5$ samples $4$ times).}
  \label{fig:cwalk-schedule-bi}
\end{minipage}
\end{figure}

For the IW-PGPE method,
we need to determine the schedule of collecting $20$ samples
under the fixed budget scenario.
First, we illustrate how the choice of sampling schedules affects the performance
of IW-PGPE.
Figure~\ref{fig:cwalk-schedule-pure} and Figure~\ref{fig:cwalk-schedule-bi}
show expected returns averaged over $100$ runs
under the sampling schedule that a batch of $k$ samples are gathered $20/k$ times
for different $k$ values.
In our implementation of IW-PGPE,
policy update is performed $100$ times after observing each batch of $k$ samples,
because we empirically observed that this performs
better than performing policy update only once.
Figure~\ref{fig:cwalk-schedule-pure} shows that
the performance of IW-PGPE depends heavily on the sampling schedule,
and gathering $k=20$ samples at once is shown to be the best choice in the Gaussian case.
Figure~\ref{fig:cwalk-schedule-bi} shows that
gathering $k=20$ samples at once is also the best choice in the bimodal case.

Although the best sampling schedule is not accessible in practice,
we use this optimal sampling schedule for IW-PGPE.
Figure~\ref{fig:cwalk-results-pure} and Figure~\ref{fig:cwalk-results-bi}
also include returns of IW-PGPE averaged over $100$ runs
as functions of the sampling steps.
These graphs show that IW-PGPE can improve the policies only in the beginning,
because all samples are gathered at once in the beginning.
The performance of IW-PGPE may be further improved
if it is possible to gather more samples,
but this is prohibited under the fixed budget scenario.
On the other hand, return values of M-PGPE constantly increase throughout iterations,
because artificial samples can be kept generated without additional sampling costs.
This illustrates a potential advantage of model-based RL methods.

\subsection{Humanoid Robot Control}
\label{subsec:humanoid}
\begin{figure}[t]
\centering
	\subfigure[CB-i]{
          ~~~~\includegraphics[clip, height=15em]{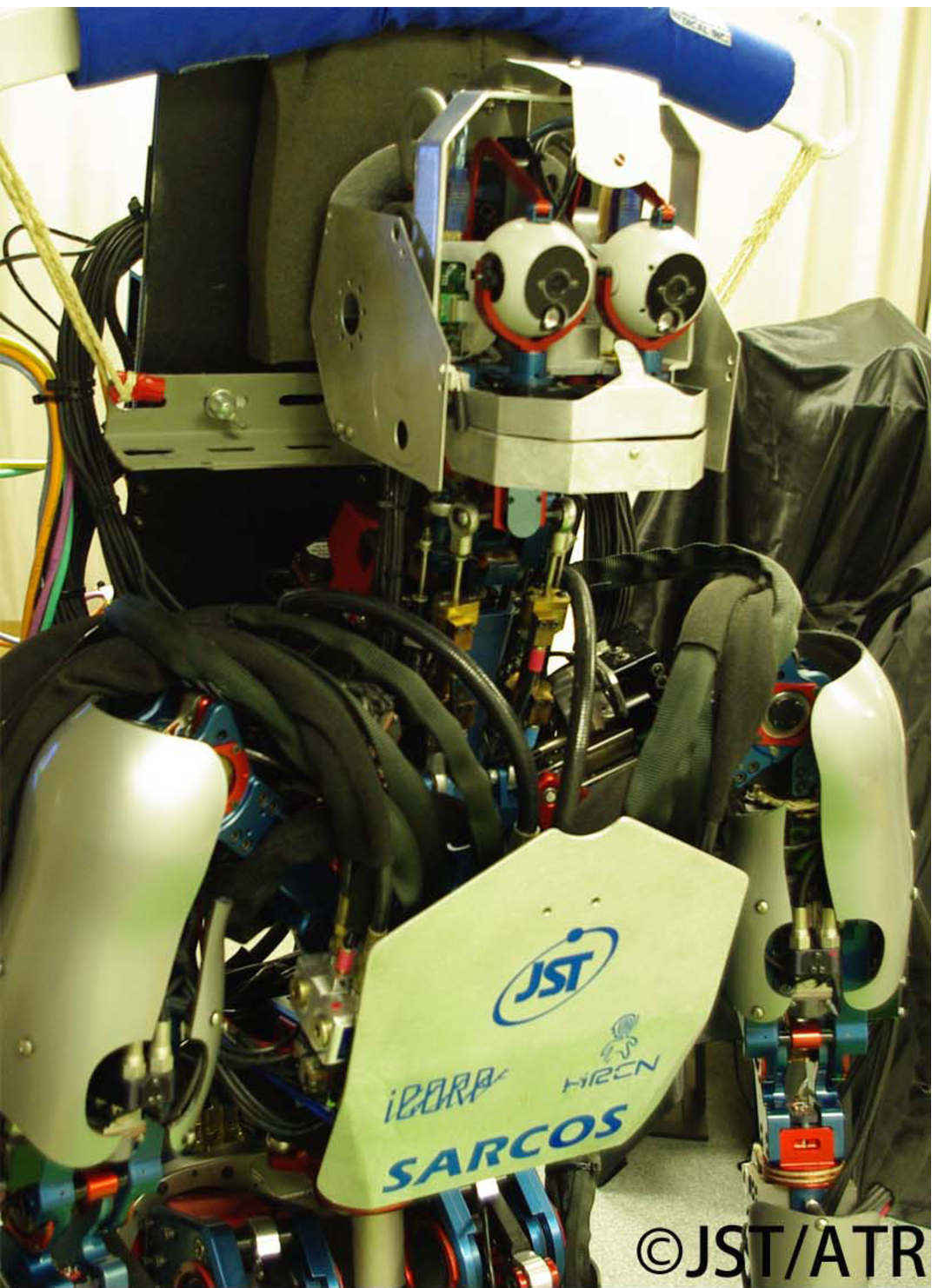}~~~~
          \label{fig:cbi}}
	\subfigure[Simulator of the CB-i upper-body]{
          ~~~~\includegraphics[clip, height=15em]{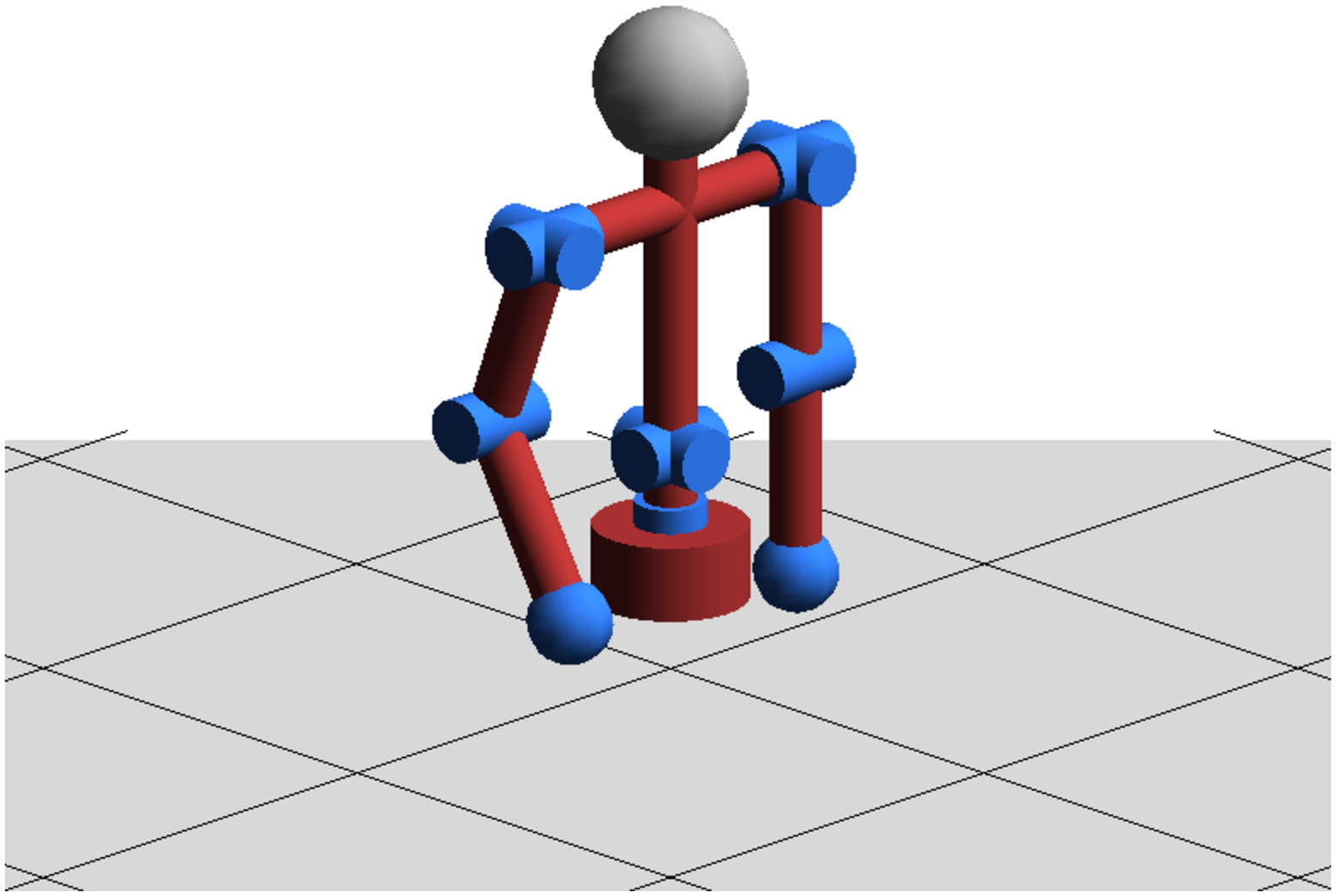}~~~~
          \label{fig:simulator}}
	\caption{Humanoid robot CB-i and its upper-body model.}
	\label{fig:CBI}
\end{figure}

Finally, we evaluate the performance of M-PGPE on a practical control problem of a simulated upper-body model of the humanoid robot \emph{CB-i} \cite{Cheng:2007} (see Figure~\ref{fig:cbi}). We use its simulator for experiments (see Figure~\ref{fig:simulator}).
The goal of the control problem is to lead the end-effector of the right arm
(right hand) to the target object.

\subsubsection{Setup}
The simulator is based on the upper-body of the CB-i humanoid robot, which has 9 joints for shoulder pitch, shoulder roll, elbow pitch of the right arm, shoulder pitch, shoulder roll, elbow pitch of the left arm, waist yaw, torso roll, and torso pitch.

At each time step, the controller receives a state vector from the system and sends out an action vector.
The state vector is 18-dimensional and real-valued, which corresponds to the current angle in degree and the current angular velocity for each joint.
The action vector is 9-dimensional and real-valued, which corresponds to the target angle of each joint in degree.

We simulate a noisy control system by perturbing action vectors with independent bimodal Gaussian noise.
More specifically, for each action element, we add Gaussian noise with mean $0$ and standard deviation $3$ with probability $0.6$, and Gaussian noise with mean $-5$ and standard deviation $3$ with probability $0.4$.

The initial posture of the robot is fixed to standing up straight with arms down.
The the target object is located in front-above of the right hand which is reachable by using the controllable joints. The reward function at each time step is defined as
\[
r_t=\exp(-10 d_t) - 0.000005 \min\{c_t,1000000\},
\]
where $d_t$ is the distance between the right hand and target object at time step $t$,
and $c_t$ is the sum of control costs for each joint. The coefficient $0.000005$ is multiplied to keep the values of the two terms in the same order of magnitude.
The deterministic policy model used in PGPE is defined as $a=\boldsymbol{\theta}^\top \boldsymbol{\phi(s)}$ with the basis function $\boldsymbol{\phi(s)} = \boldsymbol{s}$.
We set the episode length at $T = 100$, the discount factor at $\gamma = 0.9$,
and the learning rate at $\lrate=0.1/\|\nabla_{\rho} \hat{J}(\bm{\rho})\|$.

\subsubsection{Experiment with 2 Joints}
First, we only use 2 joints among the 9 joints, i.e., we allow only the right shoulder pitch and right elbow pitch to be controlled, while the other joints remain still at each time step (no control signal is sent to these joints). Therefore, the dimensionality of state vector $\boldsymbol{s}$
and action vector $\boldsymbol{a}$ is $4$ and $2$, respectively.
Under this simplified setup, we compare the performance of M-PGPE(LSCDE), M-PGPE(GP), and IW-PGPE.

\begin{figure}[p]
  \centering
\includegraphics[width=0.6\textwidth,clip]{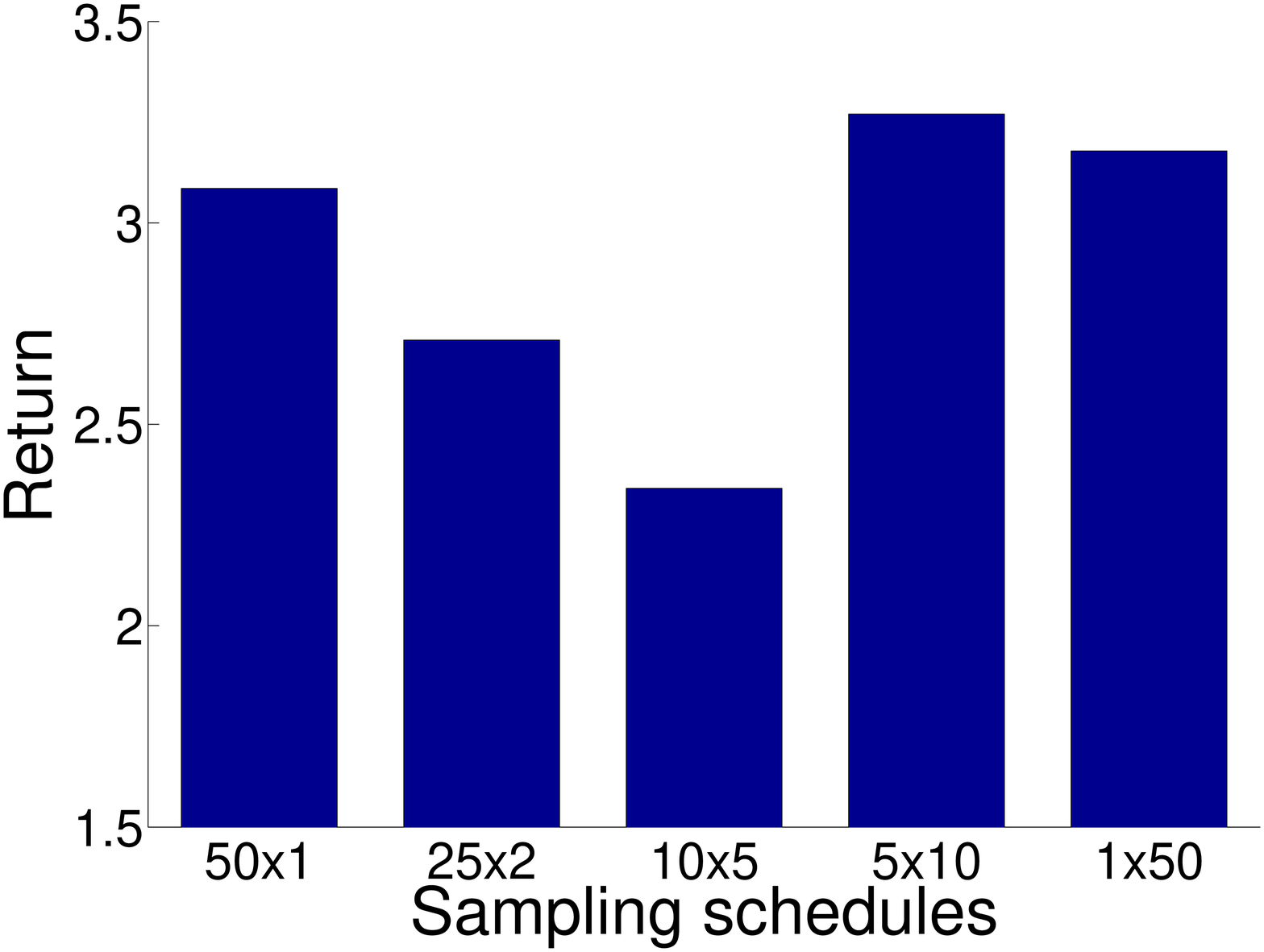}
\caption{
	Returns obtained by IW-PGPE averaged over 10 runs in 2-joint humanoid robot simulator for different sampling schedules (e.g., $5 \times 10$ means gathering $k=5$ samples 10 times).
}
\label{fig:simu_schedule2j}
\vspace*{10mm}
  \includegraphics[width=0.6\textwidth,clip]{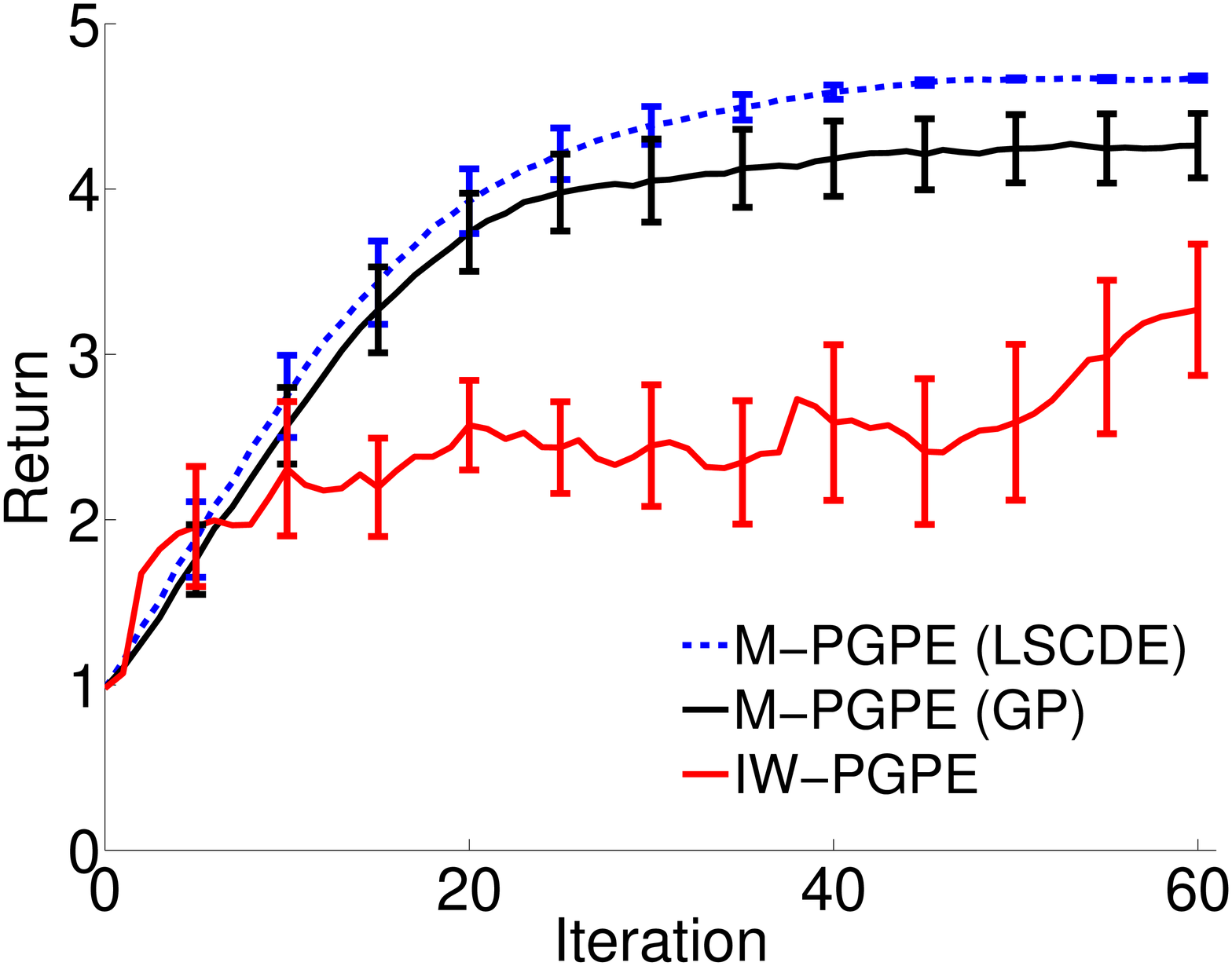}
  \caption{
  	Averages and standard errors of obtained returns over 10 runs in 2-joint humanoid robot simulator. All methods use 50 samples for policy learning. In M-PGPE(LSCDE) and M-PGPE(GP), all 50 samples are gathered in the beginning and the environment model is learned; then 2000 artificial samples are generated in each update iteration. In IW-PGPE, a batch of 5 samples are gathered for 10 iterations, which was shown to be the best sampling scheduling (see Figure~\ref{fig:simu_schedule2j}). Note that policy update is performed 100 times after observing each batch of samples, which we confirmed to perform well. The IW-PGPE curve is elongated to have the same horizontal scale as others.
}
  \label{fig:simu_result2j}
\end{figure}

We suppose that the budget for data collection is limited to $N = 50$ episodic samples.
For the M-PGPE methods, all samples are collected at first using the uniformly random initial states and policy.
More specifically, the initial state is chosen from the uniform distributions over $\mathcal{S}$.
At each time step, the $i$-th element of action vector $a_i$ is chosen from
the uniform distribution on $[s_i-5,s_i+5]$.
In total, we have 5000 transition samples for model estimation.
Then, we generate 1000 artificial samples for policy gradient estimation and
another 1000 artificial samples for baseline estimation from the learned transition model,
and update the control policy based on these artificial samples.
For the IW-PGPE method, we performed preliminary experiments to determine
the optimal sampling schedule (Figure~\ref{fig:simu_schedule2j}),
showing that collecting $k=5$ samples $50/k$ times yields the highest average return.
We use this sampling schedule for performance comparison with the M-PGPE methods.

\begin{figure}[t]
\centering
	\subfigure[]{\includegraphics[clip, width=0.16\columnwidth]{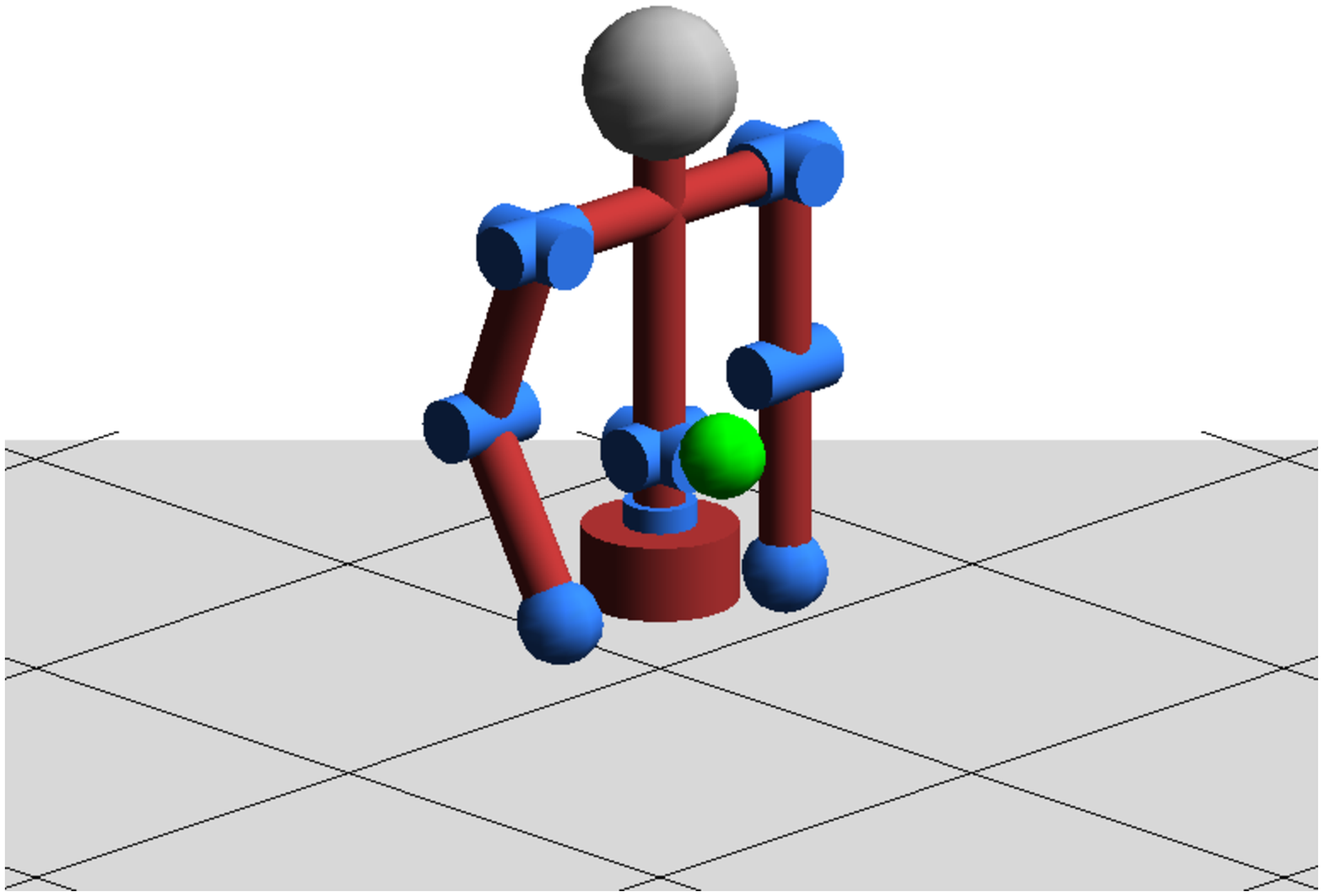}\label{fig:s0_2j}}
	\subfigure[]{\includegraphics[clip, width=0.16\columnwidth]{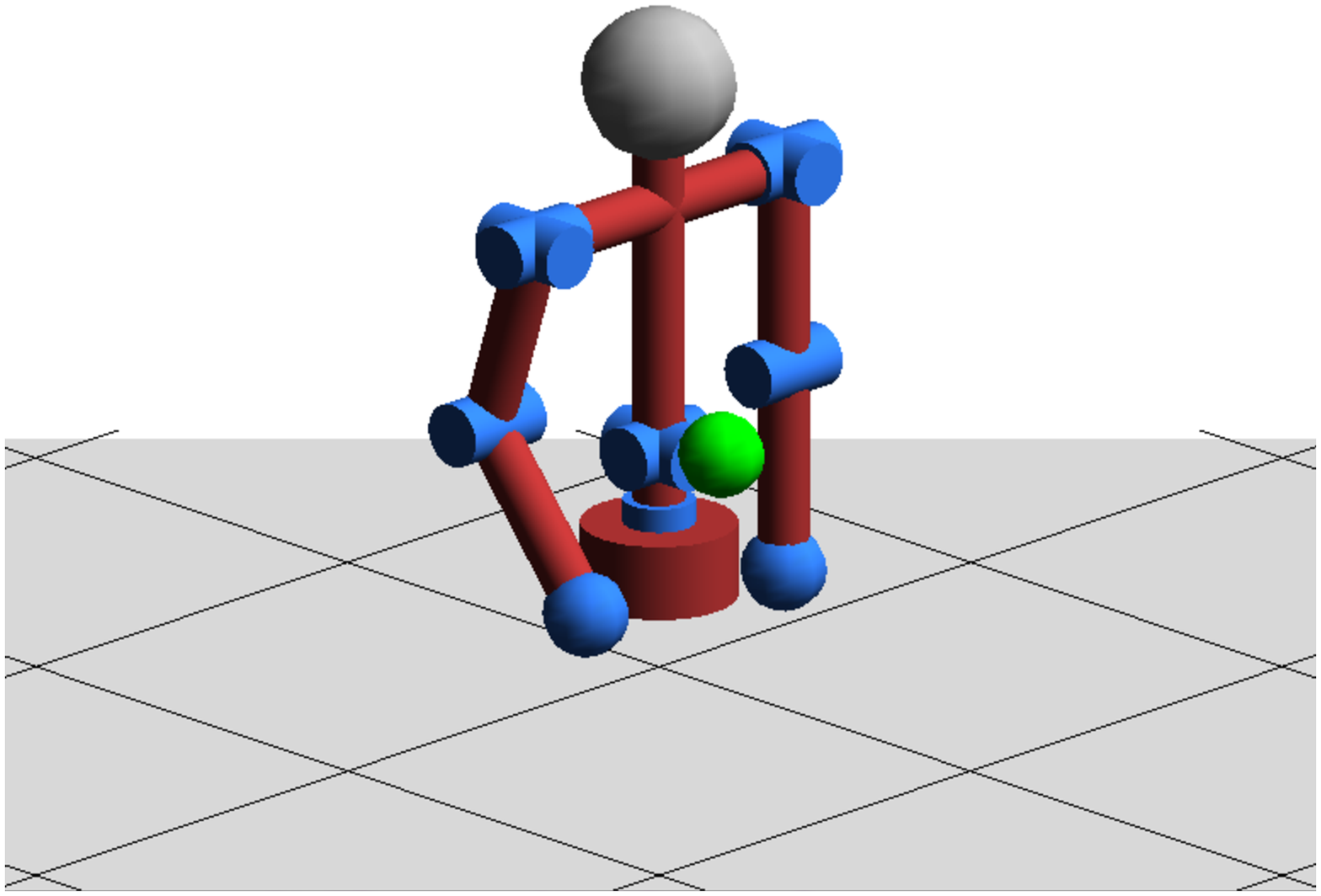}\label{fig:s2_2j}}
	\subfigure[]{\includegraphics[clip, width=0.16\columnwidth]{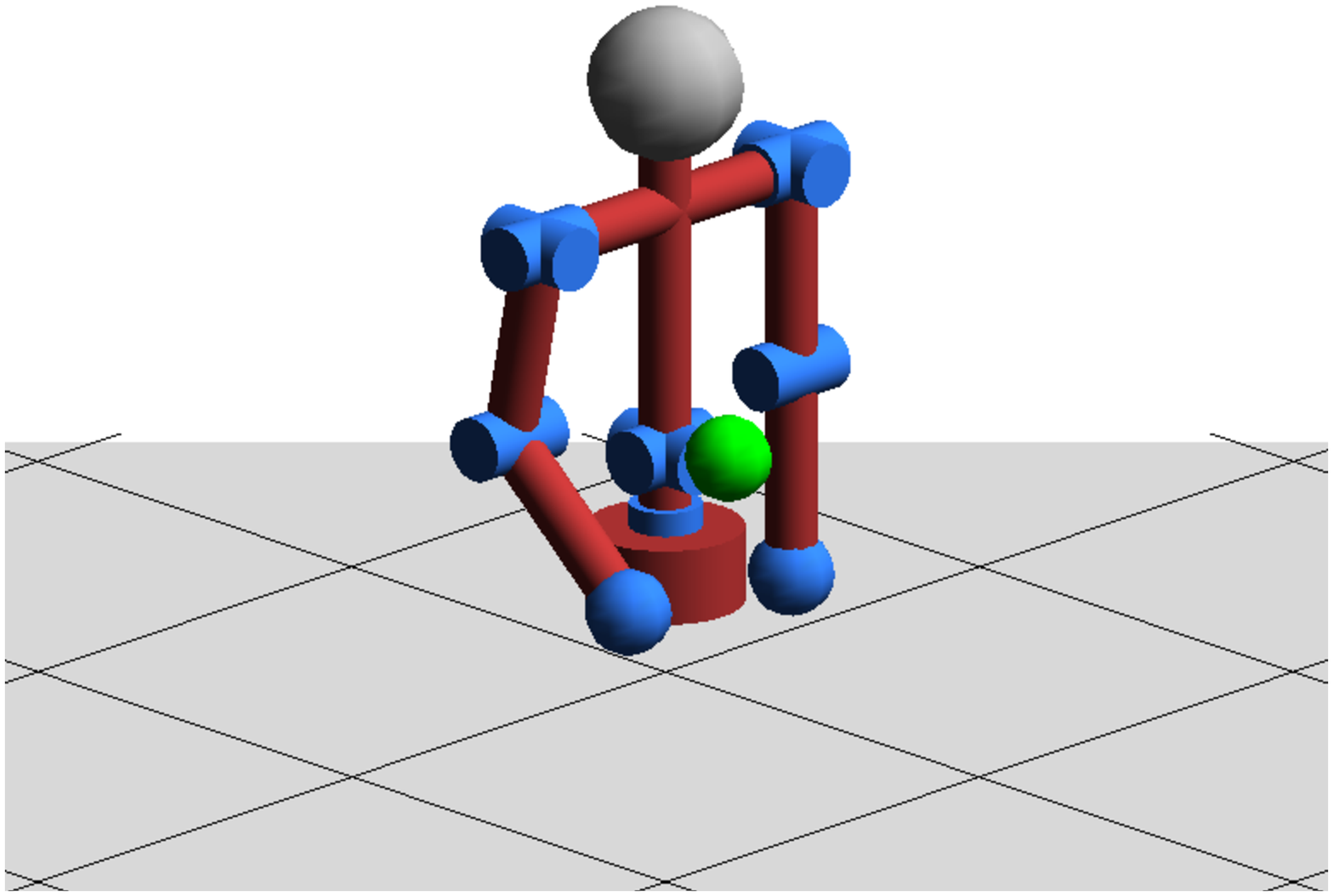}\label{fig:s4_2j}}
	\subfigure[]{\includegraphics[clip, width=0.16\columnwidth]{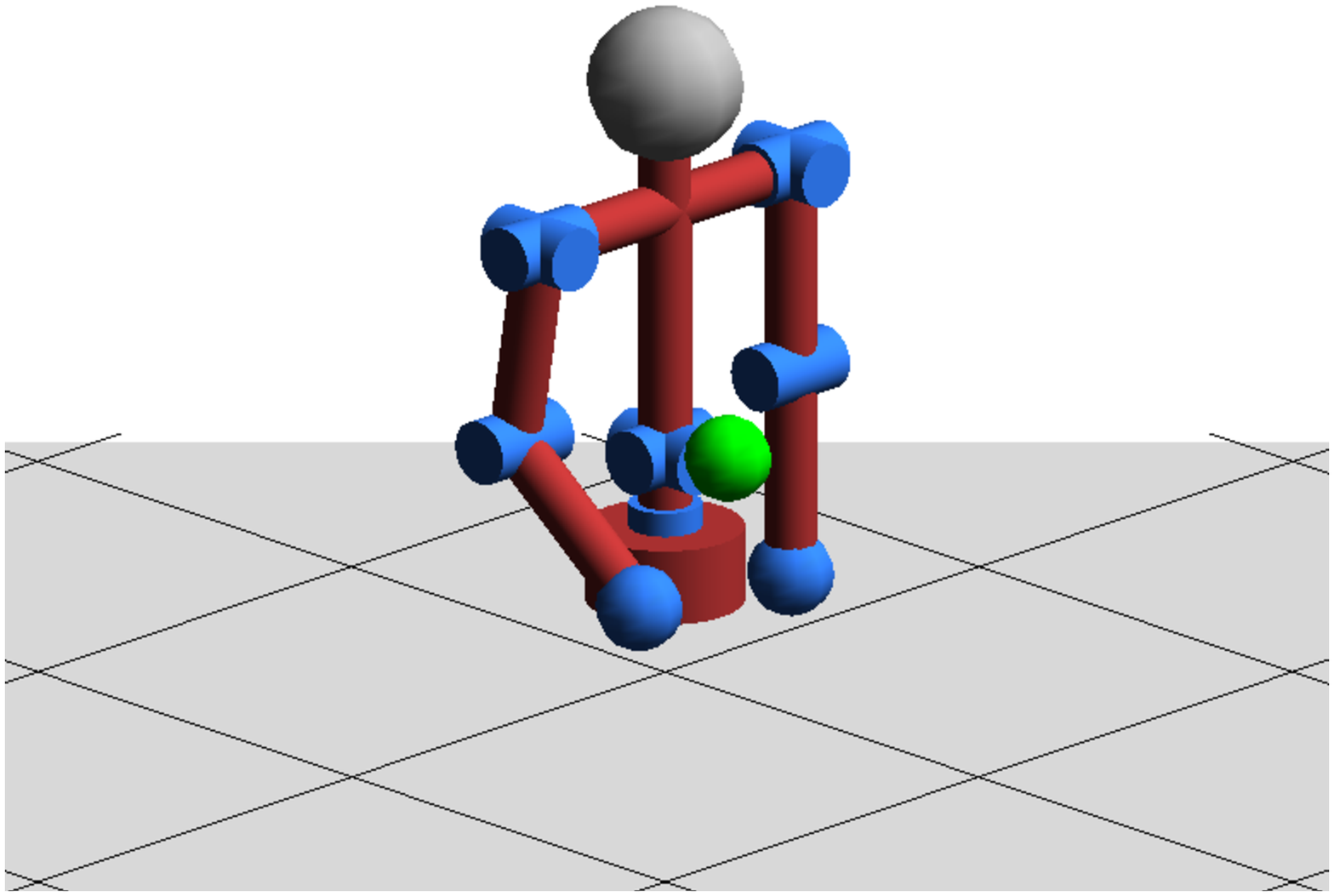}\label{fig:s5_2j}}
	\subfigure[]{\includegraphics[clip, width=0.16\columnwidth]{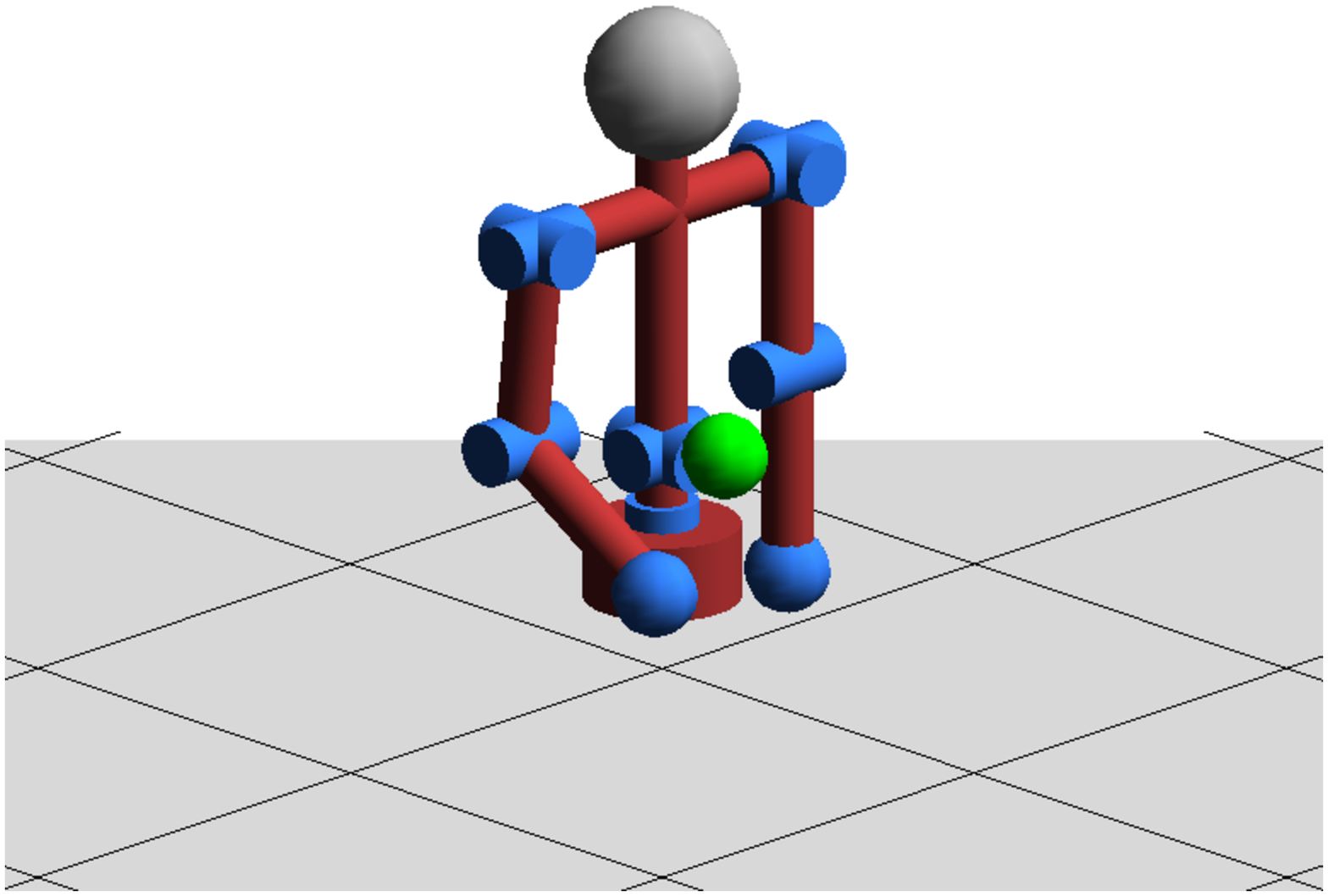}\label{fig:s6_2j}}
	\subfigure[]{\includegraphics[clip, width=0.16\columnwidth]{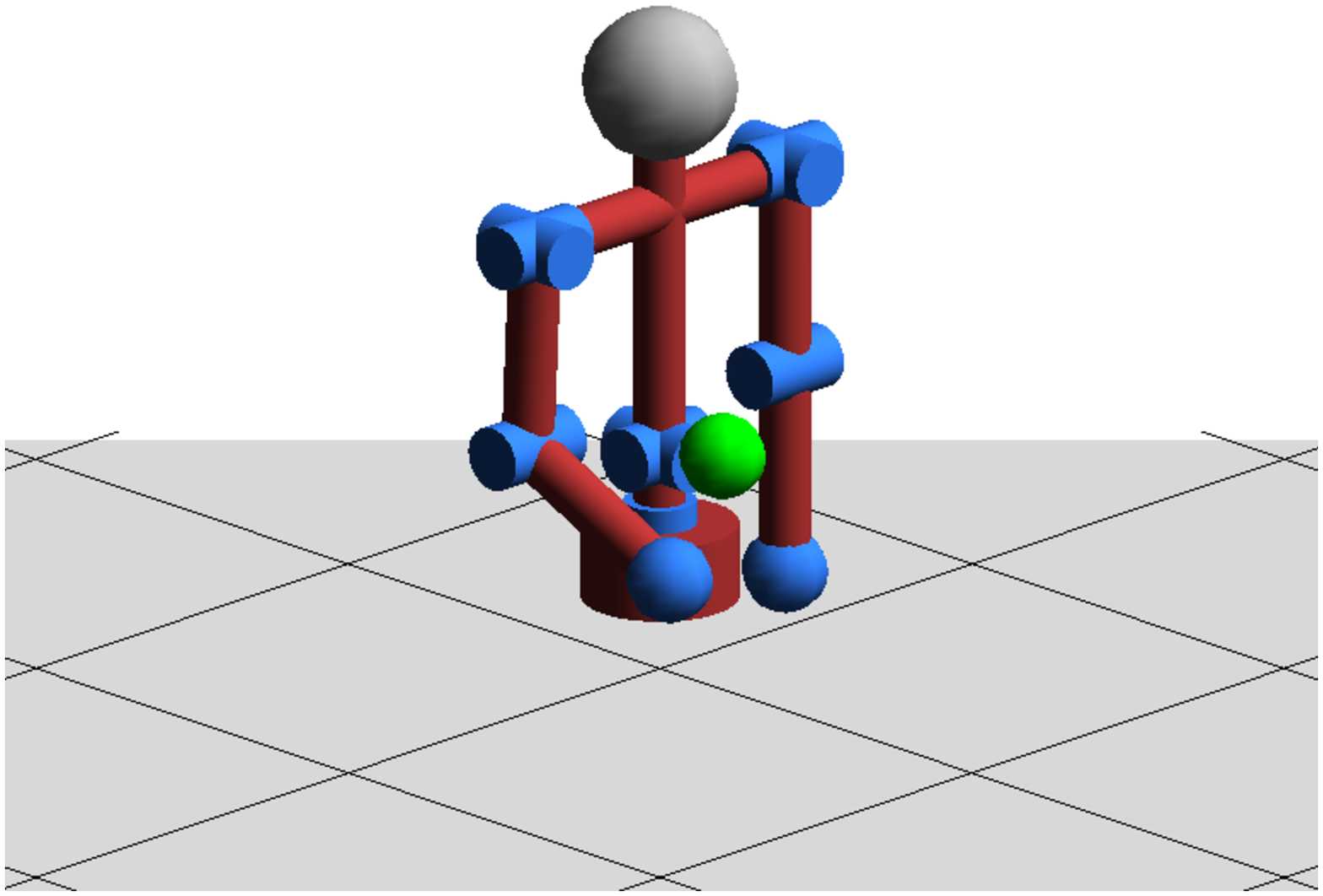}\label{fig:s7_2j}}
	\subfigure[]{\includegraphics[clip, width=0.16\columnwidth]{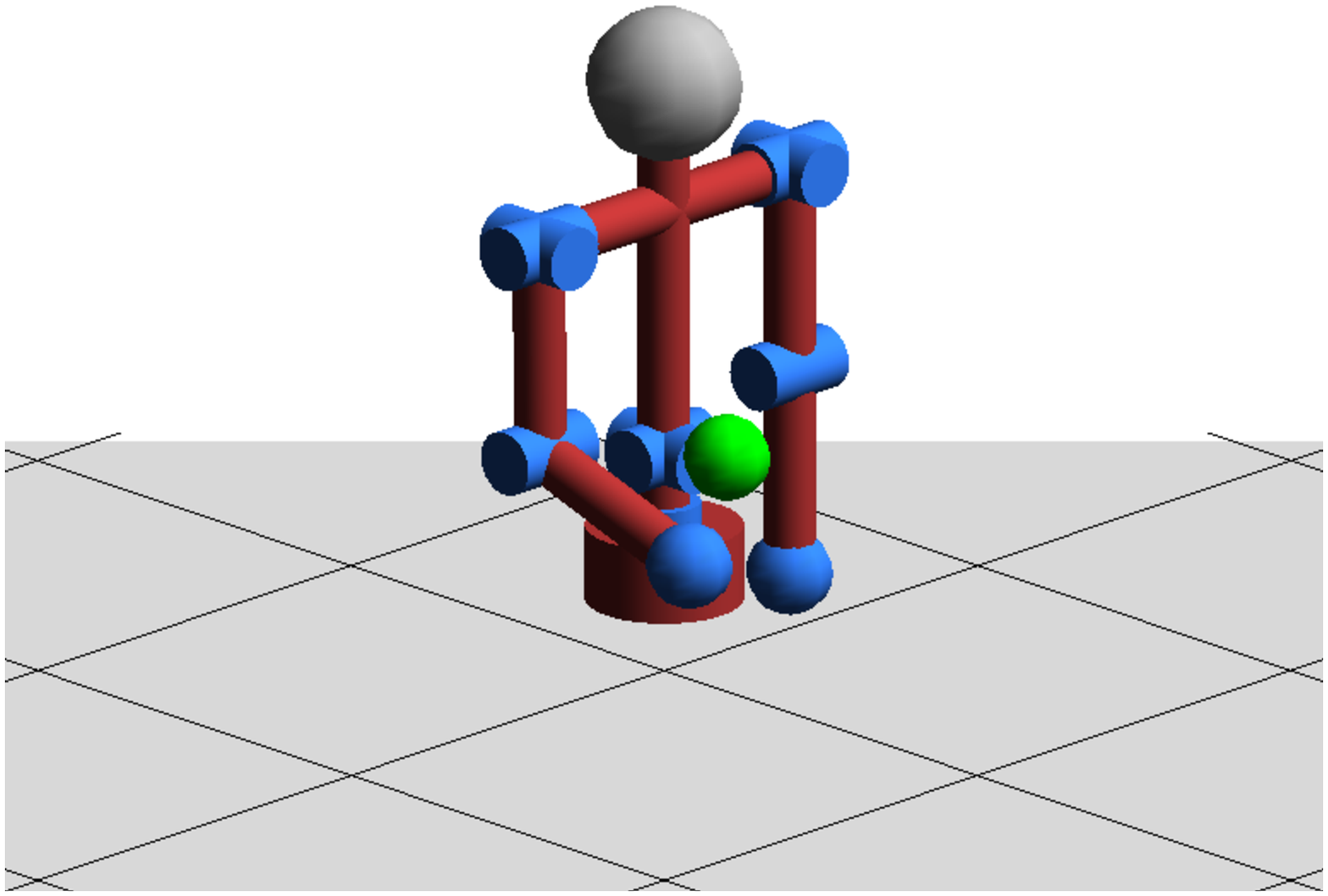}\label{fig:s8_2j}}
	\subfigure[]{\includegraphics[clip, width=0.16\columnwidth]{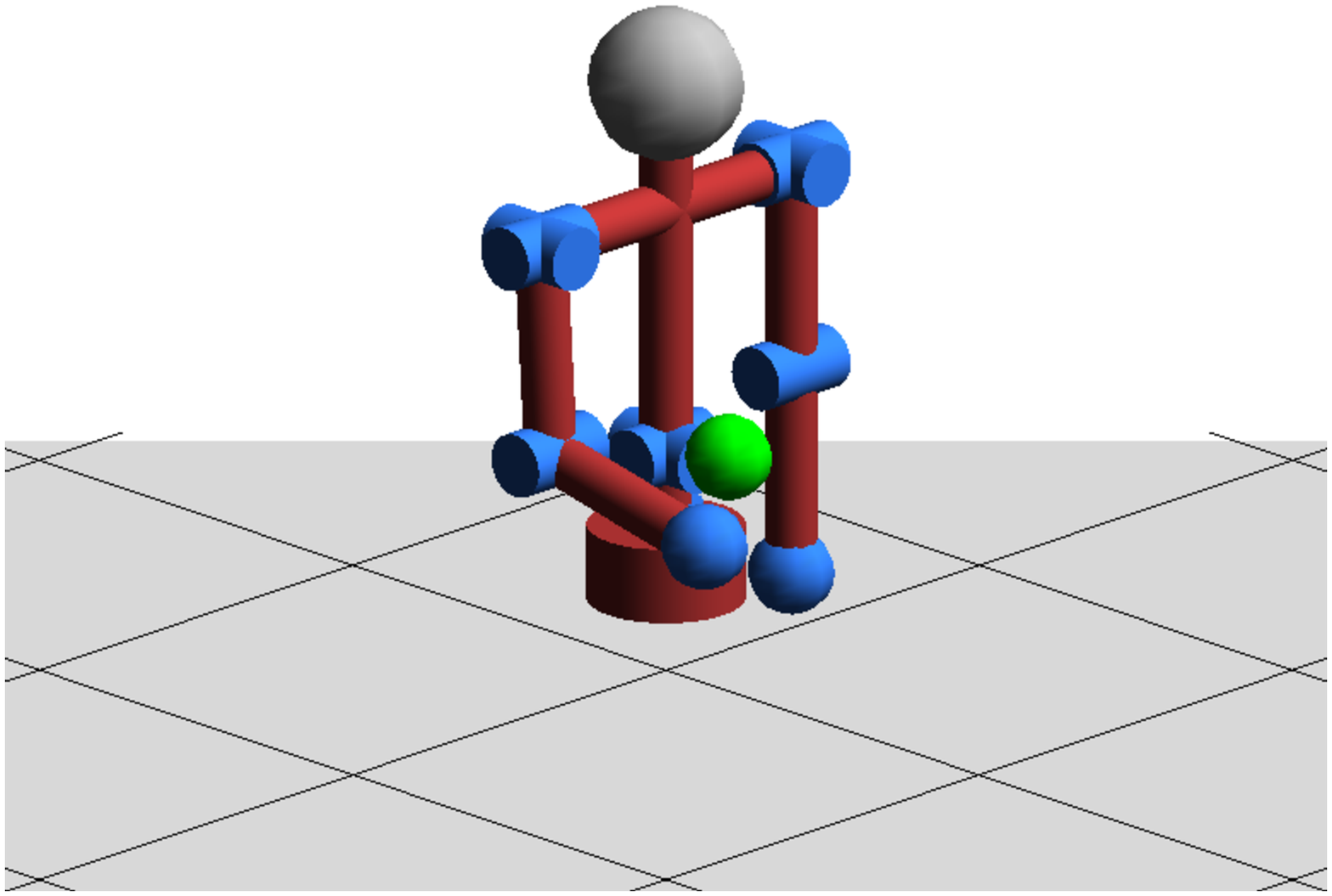}\label{fig:s9_2j}}
	\subfigure[]{\includegraphics[clip, width=0.16\columnwidth]{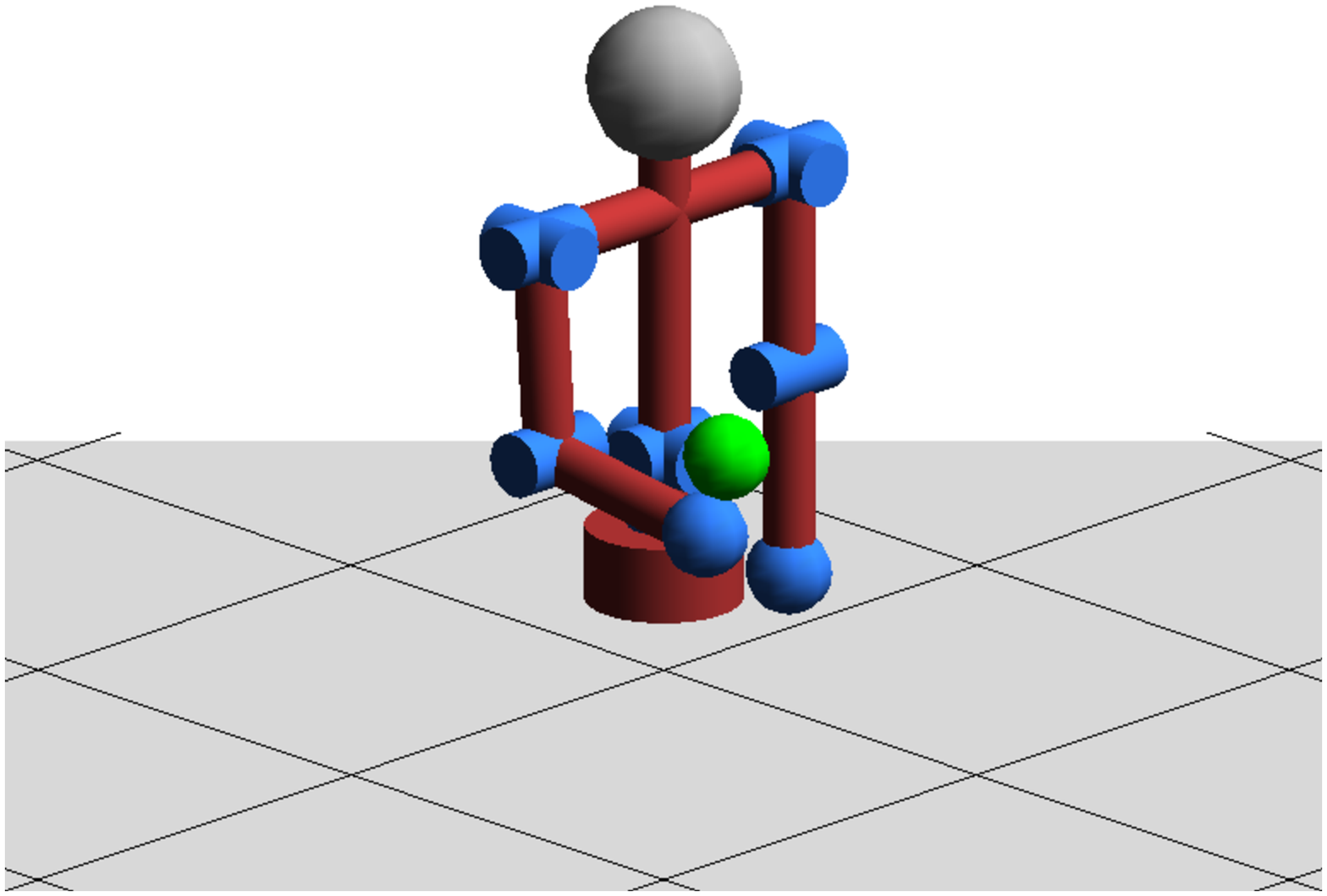}\label{fig:s10_2j}}
	\subfigure[]{\includegraphics[clip, width=0.16\columnwidth]{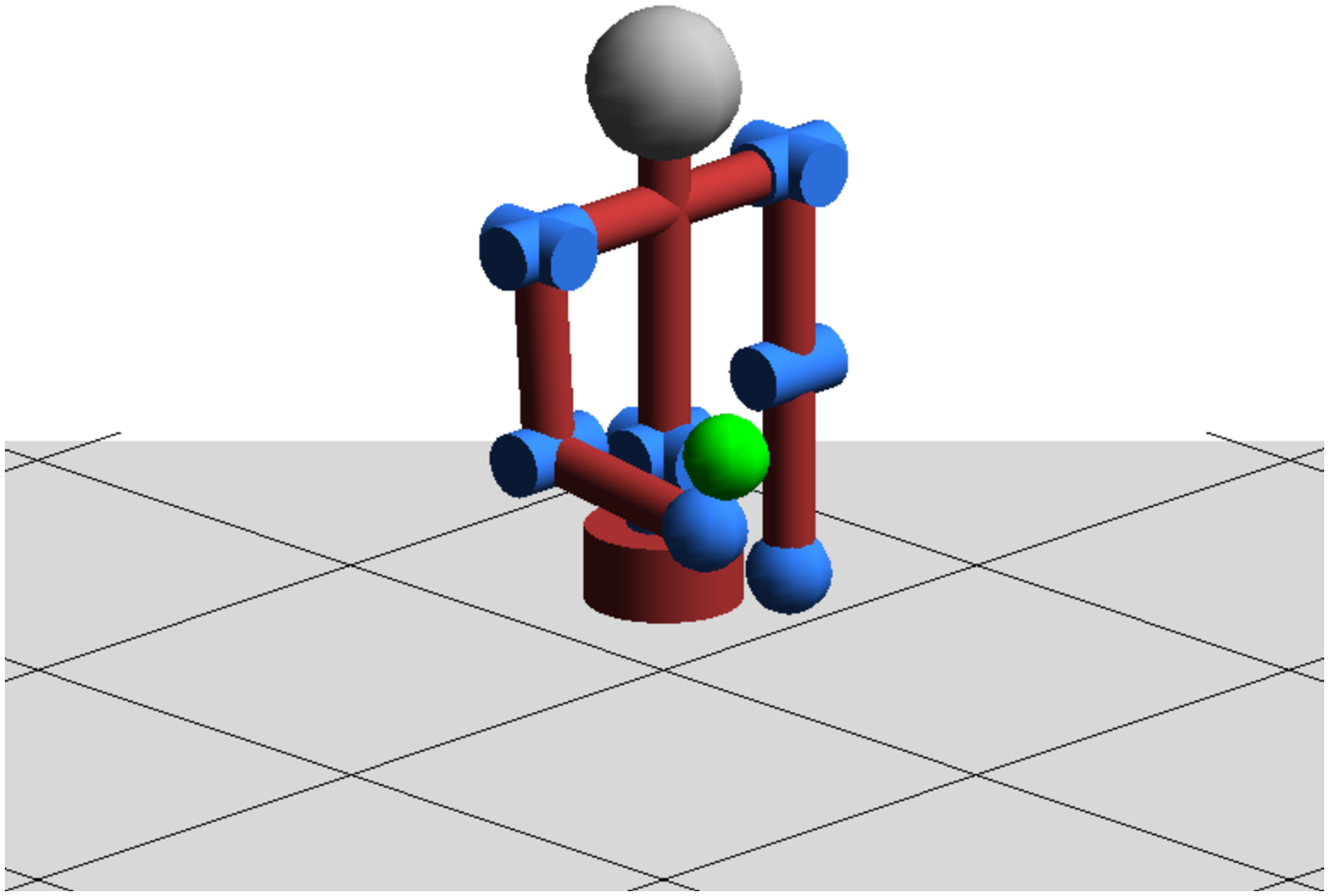}\label{fig:s11_2j}}
	\subfigure[]{\includegraphics[clip, width=0.16\columnwidth]{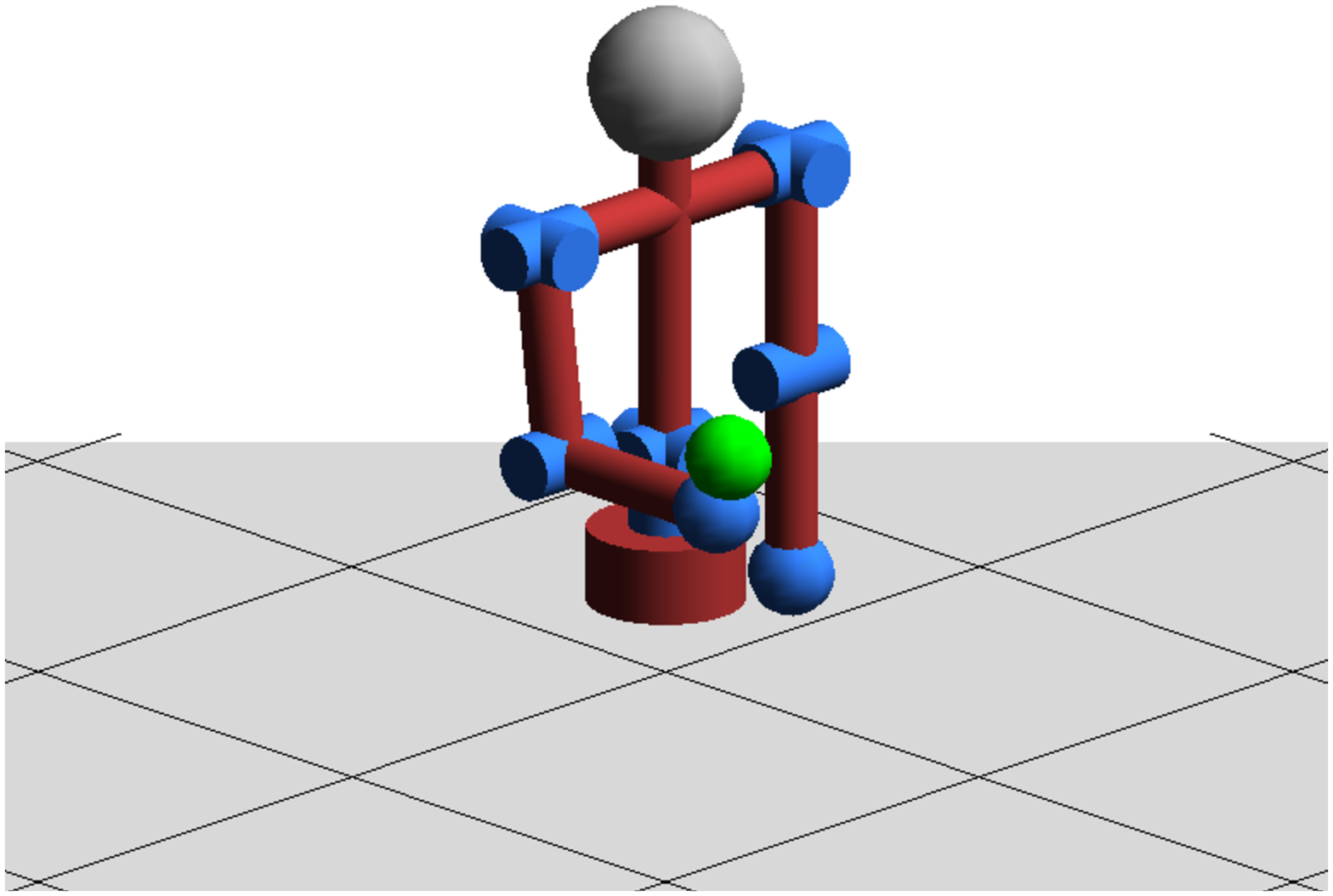}\label{fig:s12_2j}}
	\subfigure[]{\includegraphics[clip, width=0.16\columnwidth]{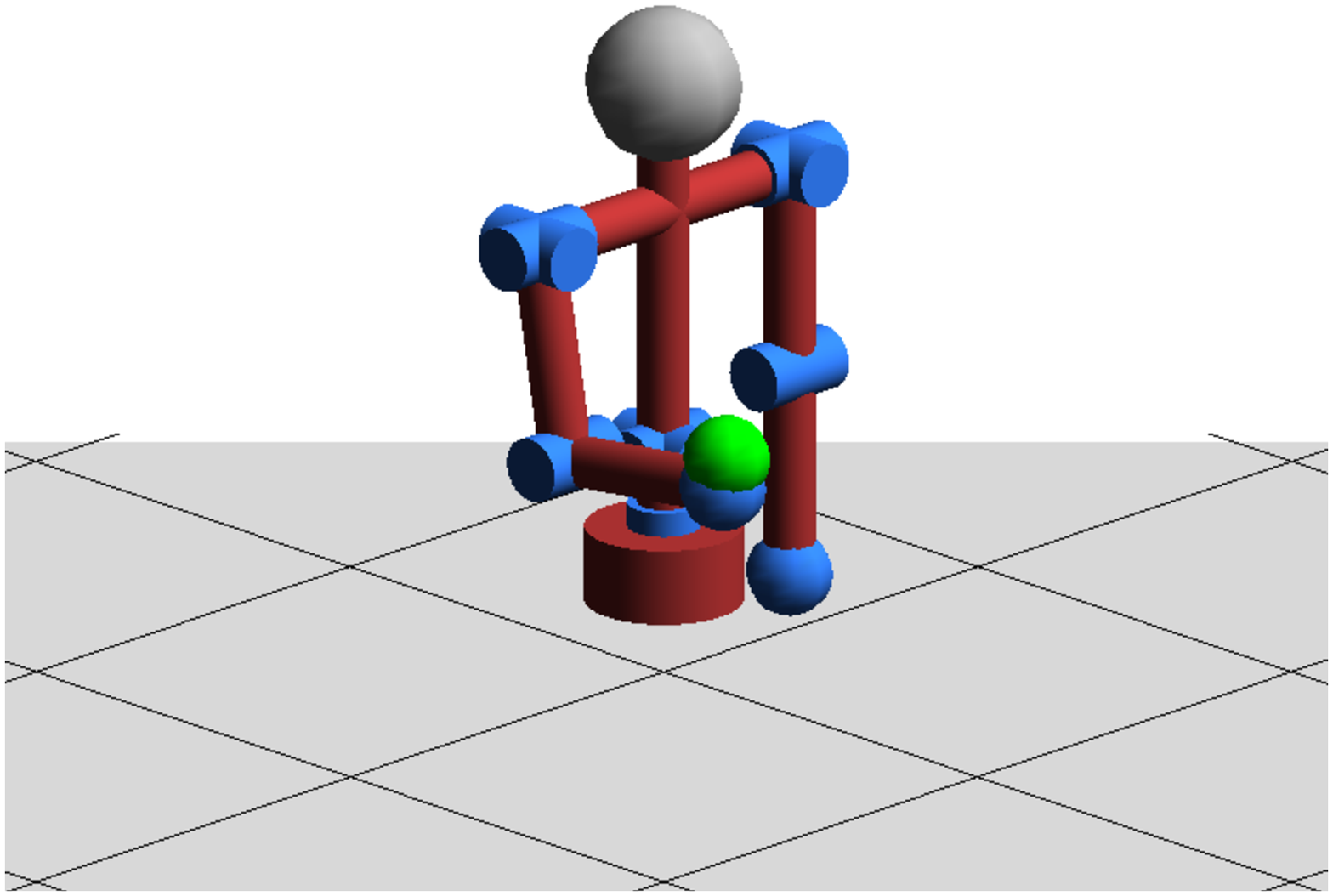}\label{fig:s13_2j}}
\caption{Example of arm reaching with $2$ joints using a policy obtained by M-PGPE at the $60$th iteration (some intermediate steps are not shown here).}
\label{fig:simu_traject2j}
\end{figure}

Returns obtained by each method averaged over 10 runs are plotted in Figure~\ref{fig:simu_result2j},
showing that M-PGPE(LSCDE) tends to outperform both M-PGPE(GP) and IW-PGPE.
Figure~\ref{fig:simu_traject2j} illustrates an example of the reaching motion
with 2-joints obtained by M-PGPE(LSCDE) at the $60$th iteration policy.
This shows that the learned policy successfully leads the right hand
to the target object within only $13$ steps in this noisy control system.

\subsubsection{Experiment with 9 Joints}
Finally, we evaluate the performance of the proposed method
on the reaching task with 9 joints, i.e., all joints are allowed to move.
In this experiment, we compare learning performance between M-PGPE(LSCDE) and IW-PGPE.
We do not include M-PGPE(GP) since it is outperformed by M-PGPE(LSCDE) on the previous 2-joints experiments,
and furthermore the GP-based method requires an enormous amount of computation time.

\begin{figure}[t]
\centering
\includegraphics[width=0.6\textwidth,clip]{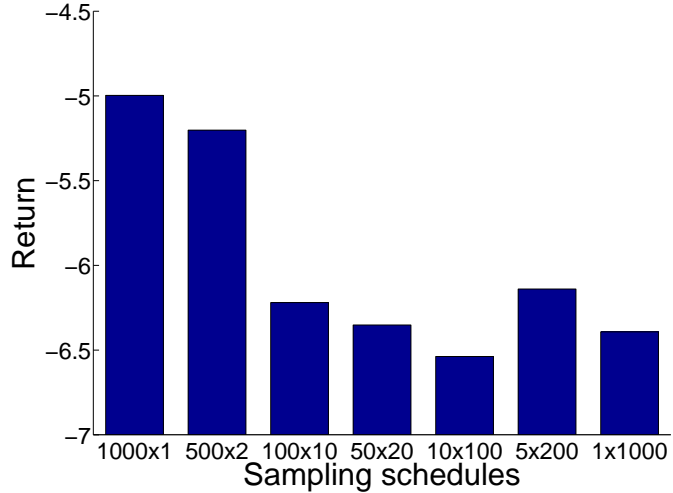}
\caption{
	Returns obtained by IW-PGPE averaged over 30 runs in humanoid robot simulator with 9 joints for different sampling schedules (e.g., $100 \times 10$ means gathering $k=100$ samples 10 times).
}
\label{fig:simu_schedule9j}
\end{figure}

\begin{figure}[t]
  \centering
  \includegraphics[width=0.6\textwidth,clip]{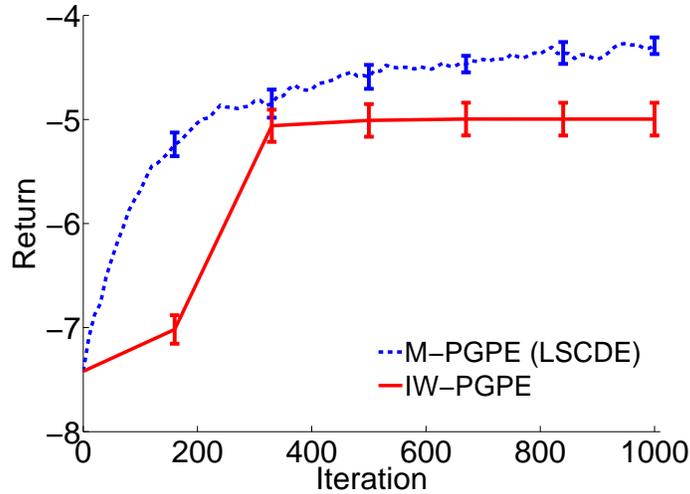}
  \caption{
	Averages and standard errors of obtained returns over 30 runs in humanoid robot simulator with 9 joints. Both methods use 1000 samples for policy learning. In M-PGPE(LSCDE), all 1000 samples are gathered in the beginning and the environment model is learned; then 2000 artificial samples are generated in each update iteration. In IW-PGPE, a batch of 1000 samples are gathered at once, which was shown to be the best scheduling (see Figure~\ref{fig:simu_schedule9j}). Note that policy update is performed 100 times after observing each batch of samples. The IW-PGPE curve is elongated to have the same horizontal scale as M-PGPE(LSCDE).
}
  \label{fig:simu_result9j}
\end{figure}

The experimental setup is essentially the same as the 2-joints experiments, but we have a budget for gathering $N = 1000$ samples for this complex and high-dimensional task.
The position of the target object is moved to far left, which is not reachable by using just 2-joints.
Thus, the robot is required to move other joints to reach the object with right hand.
  We randomly choose 5000 samples for Gaussian centers for M-PGPE(LSCDE).
  The sampling schedule for IW-PGPE was set to 1000 samples at once, which is the best sampling schedule according to Figure~\ref{fig:simu_schedule9j}. The returns obtained by M-PGPE(LSCDE) and IW-PGPE averaged over 30 runs are plotted in Figure~\ref{fig:simu_result9j}, showing that M-PGPE(LSCDE) tends to outperform the state-of-the-art IW-PGPE method in this challenging robot control task.

\begin{figure*}[t]
\centering
    \subfigure[]{\includegraphics[clip, width=0.16\columnwidth]{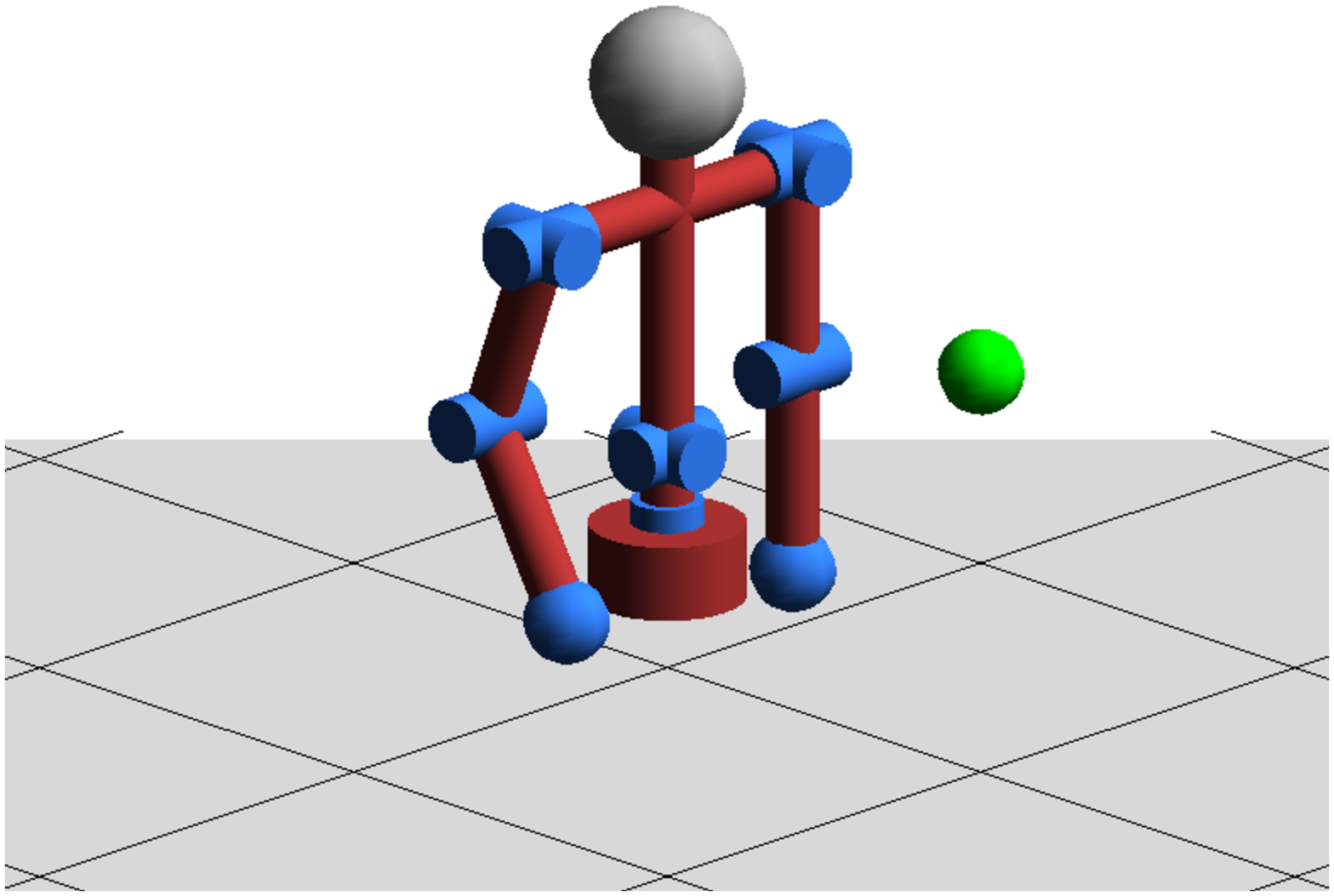}\label{fig:s0_9j}}
    \subfigure[]{\includegraphics[clip, width=0.16\columnwidth]{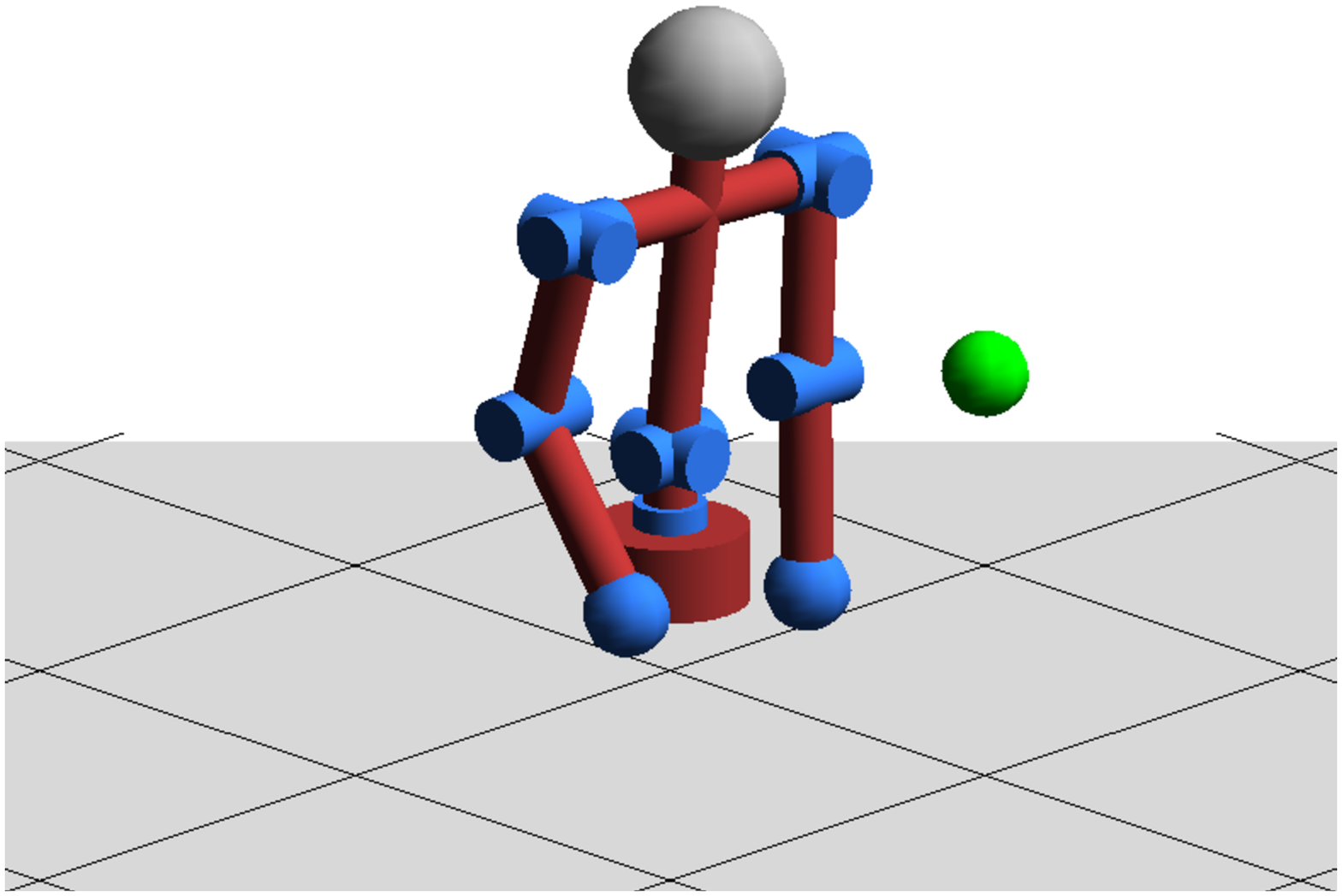}\label{fig:s2_9j}}
    \subfigure[]{\includegraphics[clip, width=0.16\columnwidth]{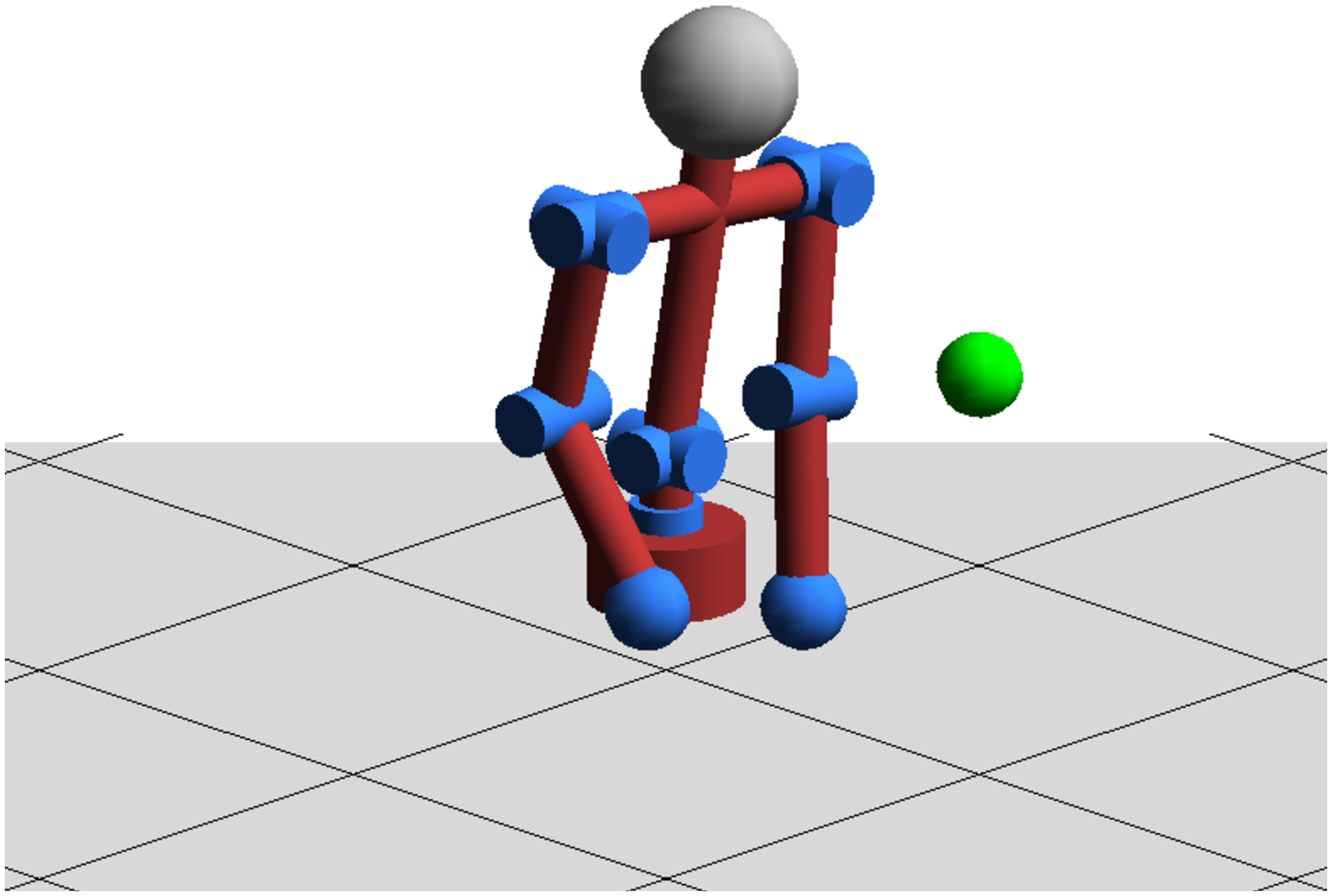}\label{fig:s3_9j}}
    \subfigure[]{\includegraphics[clip, width=0.16\columnwidth]{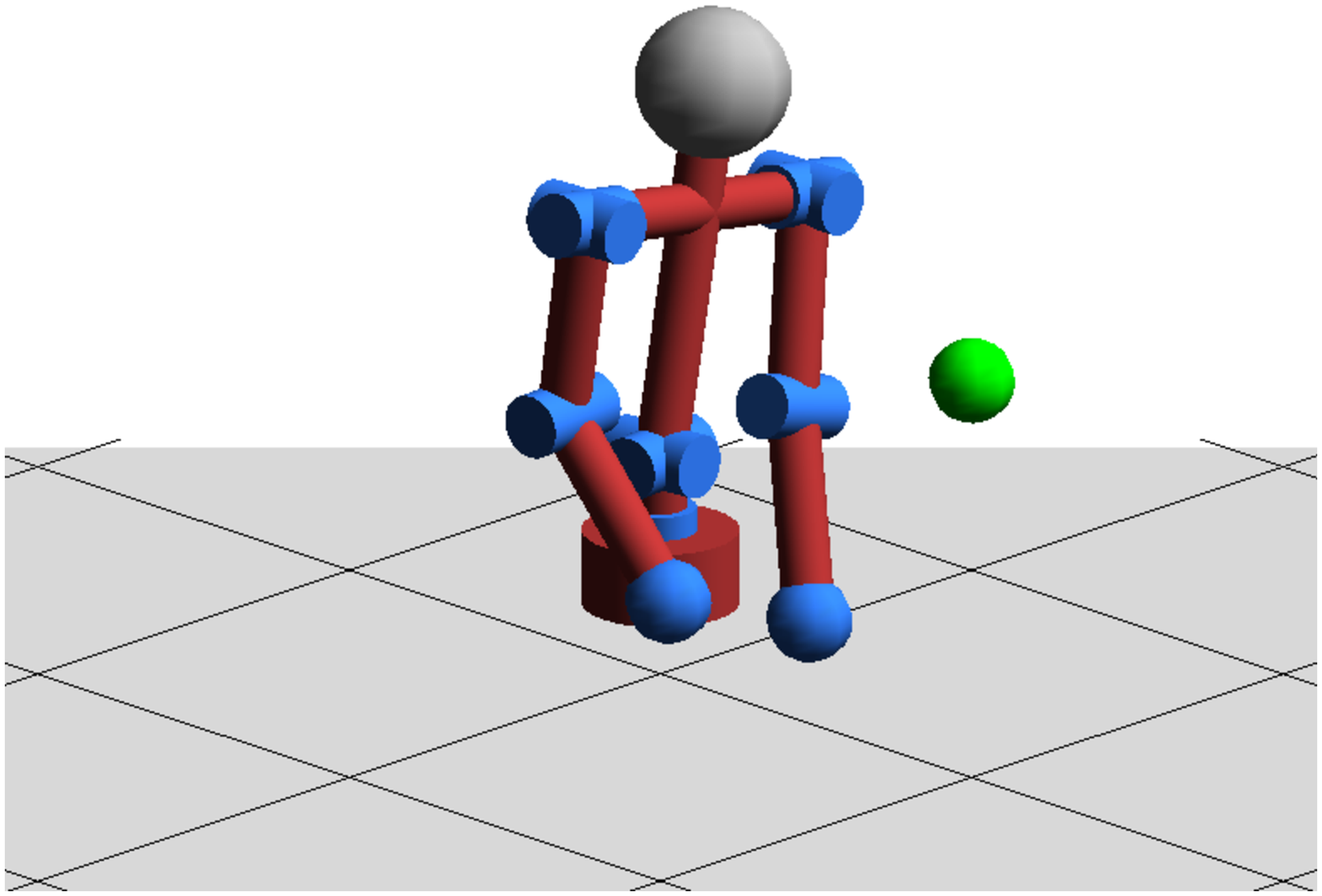}\label{fig:s4_9j}}
    \subfigure[]{\includegraphics[clip, width=0.16\columnwidth]{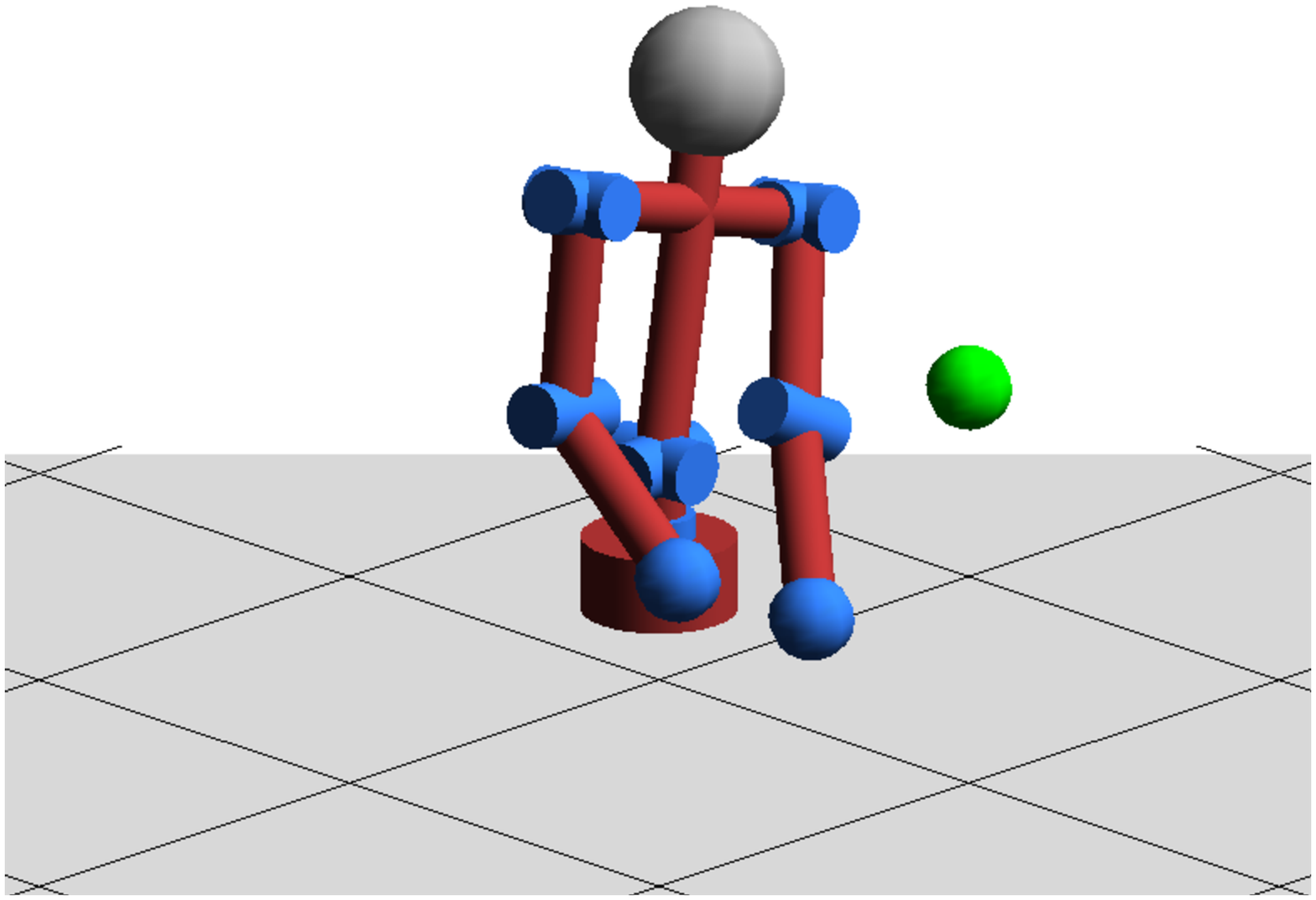}\label{fig:s6_9j}}
    \subfigure[]{\includegraphics[clip, width=0.16\columnwidth]{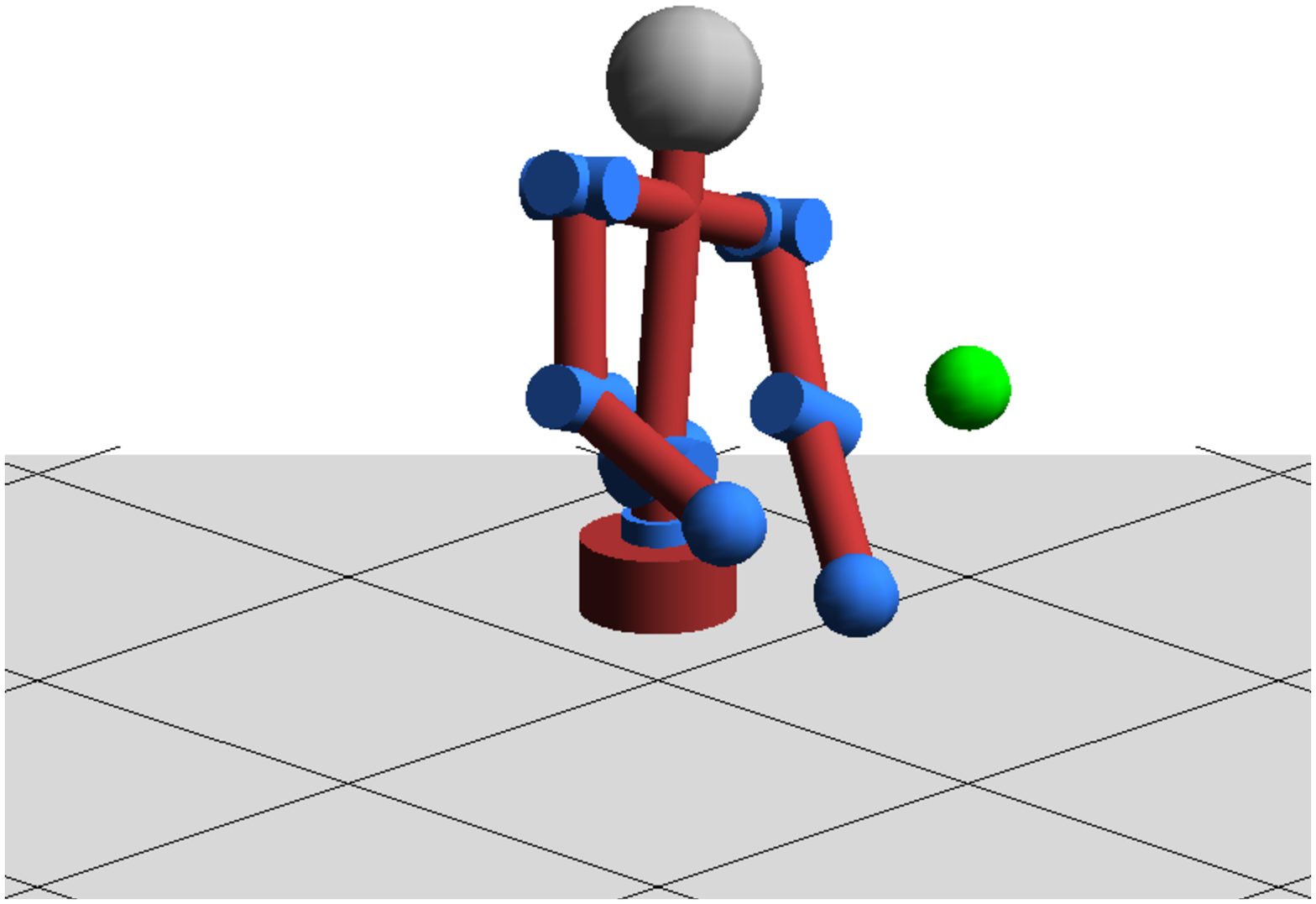}\label{fig:s8_9j}}
    \subfigure[]{\includegraphics[clip, width=0.16\columnwidth]{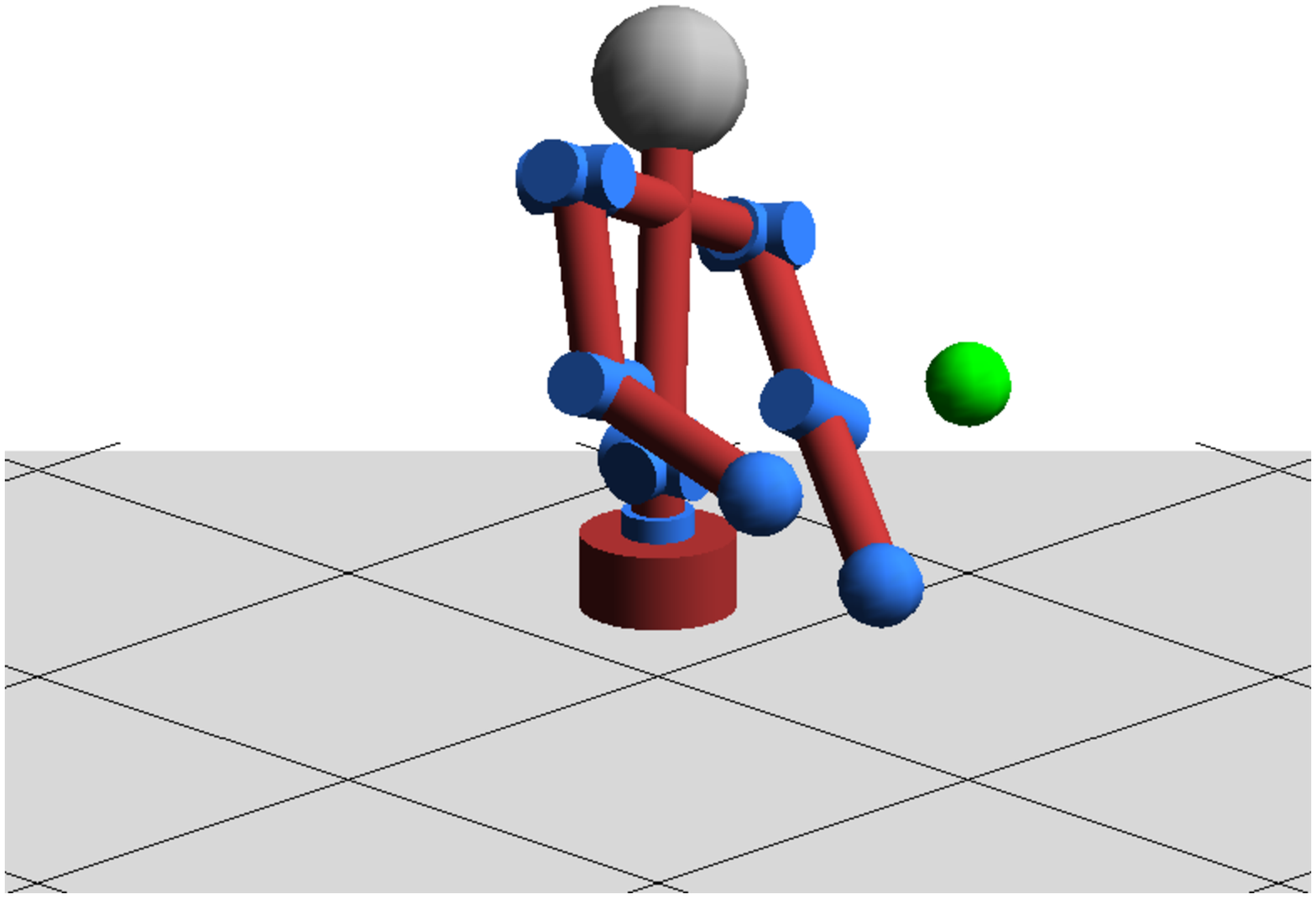}\label{fig:s9_9j}}
    \subfigure[]{\includegraphics[clip, width=0.16\columnwidth]{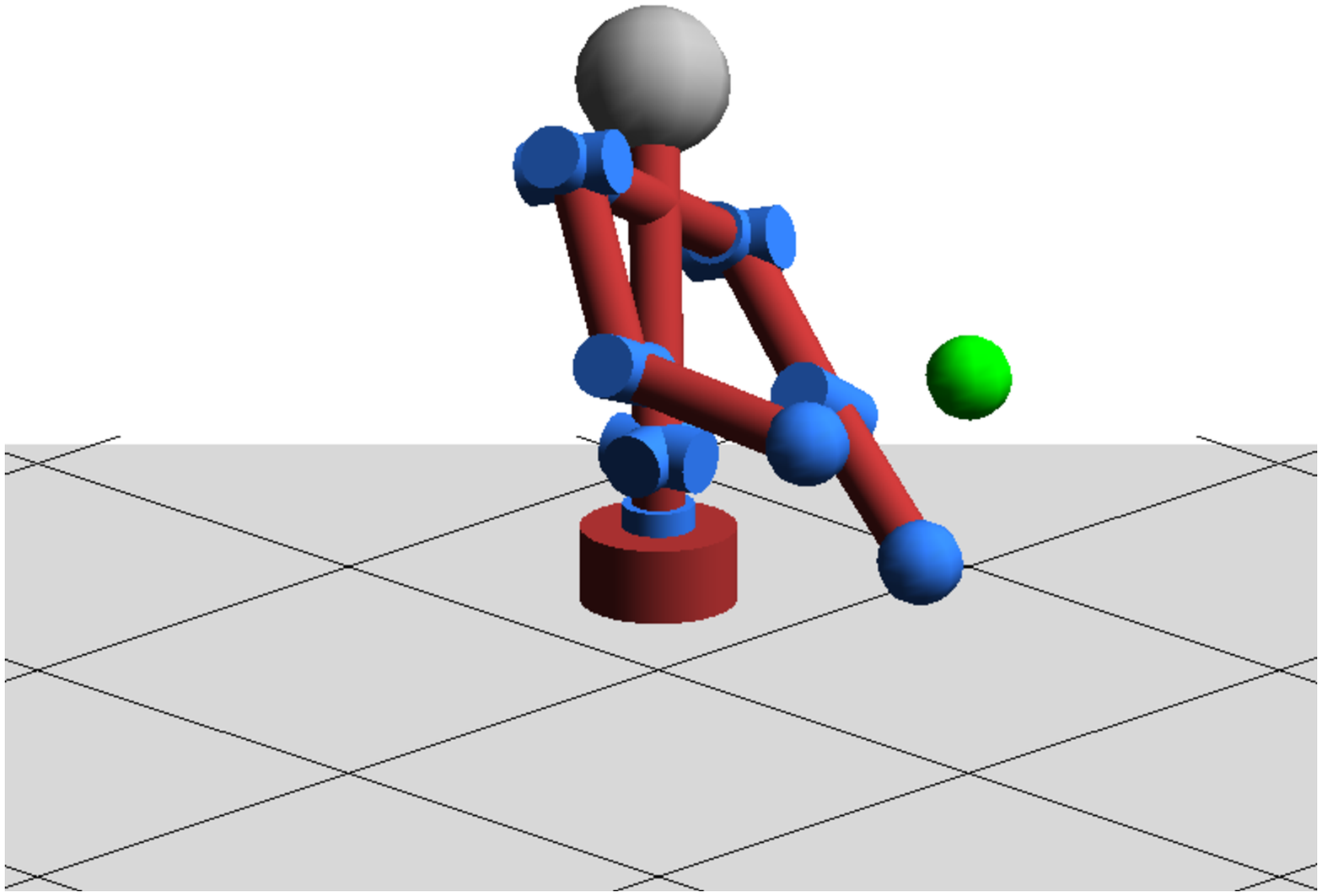}\label{fig:s10_9j}}
    \subfigure[]{\includegraphics[clip, width=0.16\columnwidth]{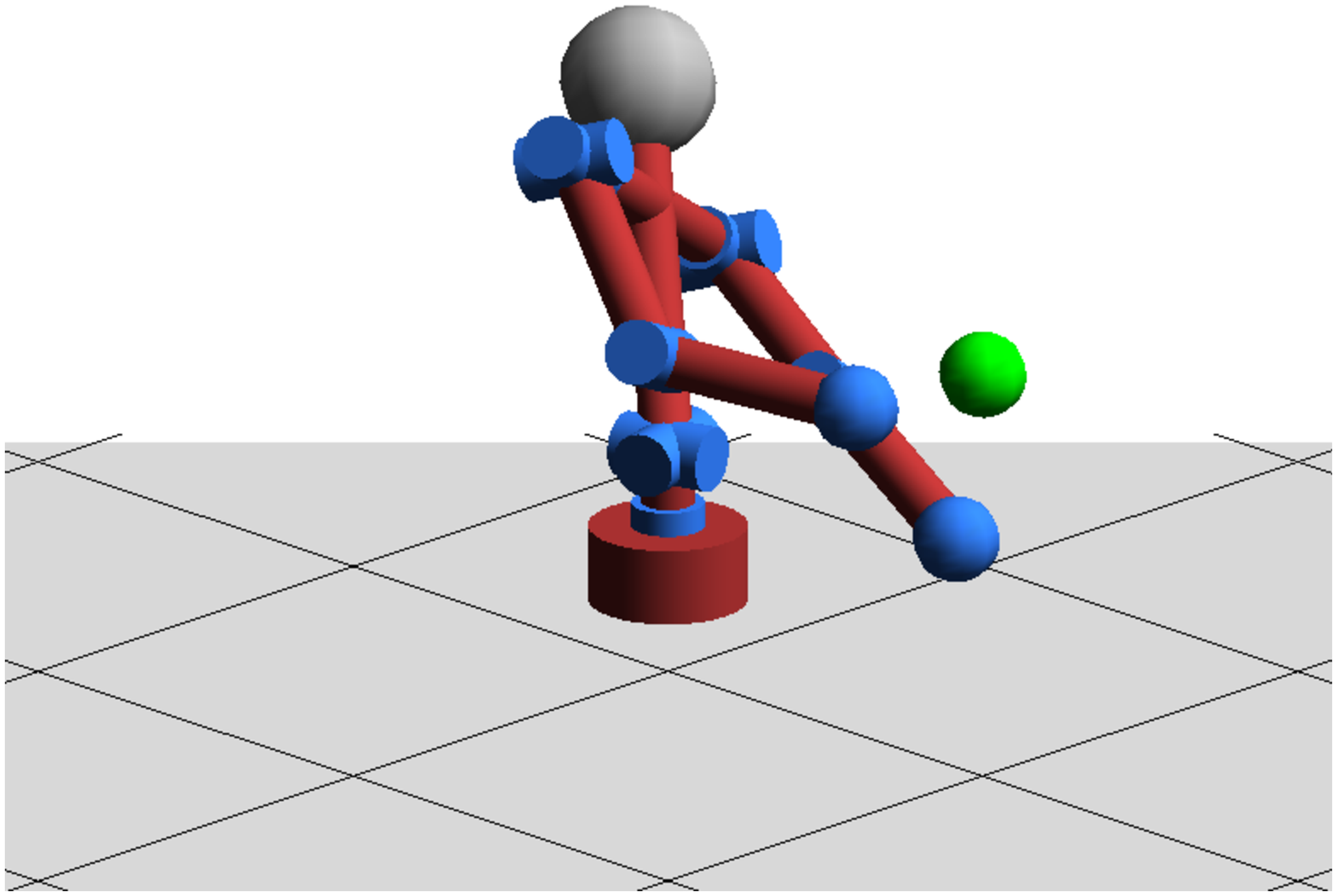}\label{fig:s11_9j}}
    \subfigure[]{\includegraphics[clip, width=0.16\columnwidth]{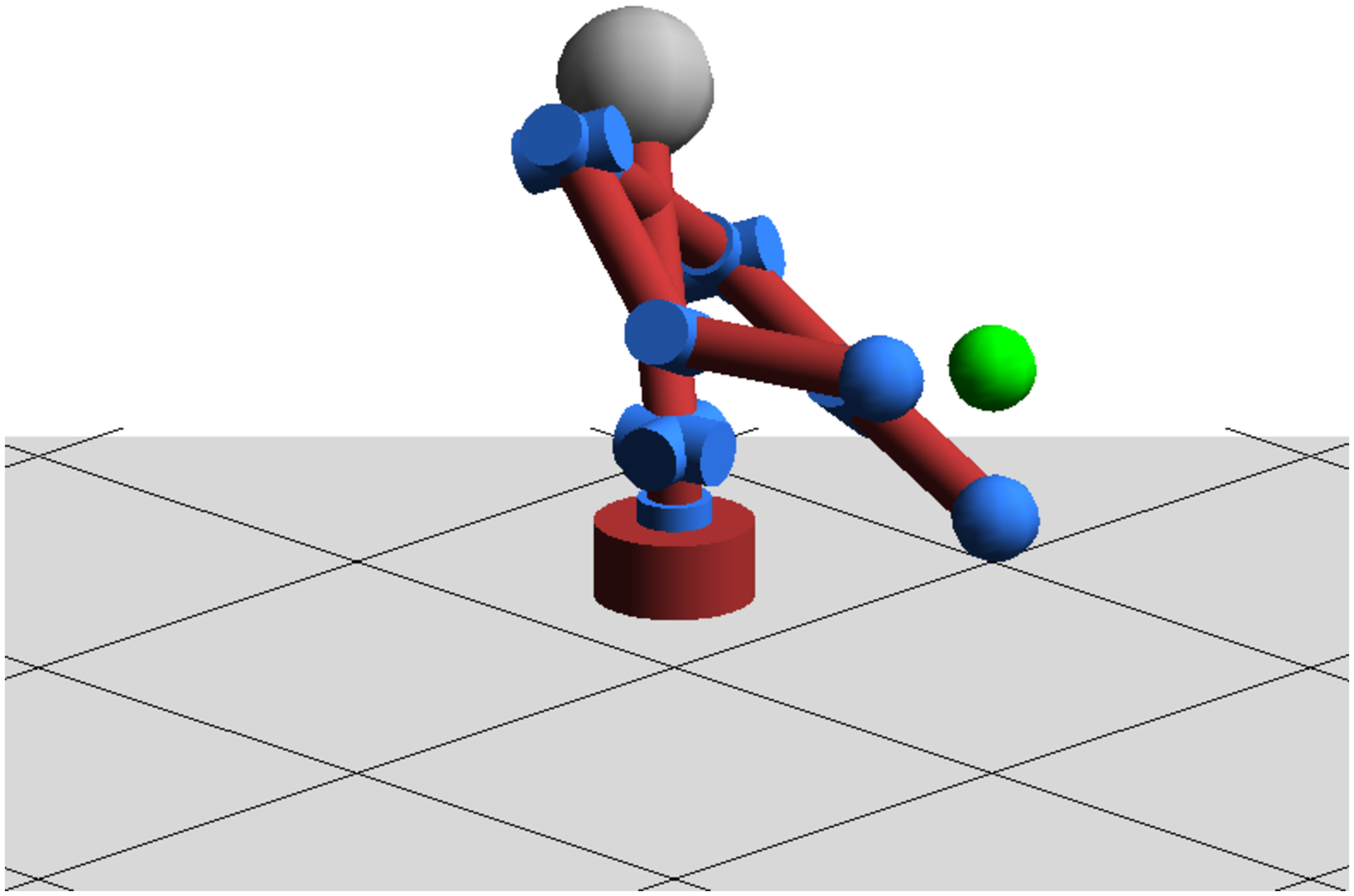}\label{fig:s12_9j}}
    \subfigure[]{\includegraphics[clip, width=0.16\columnwidth]{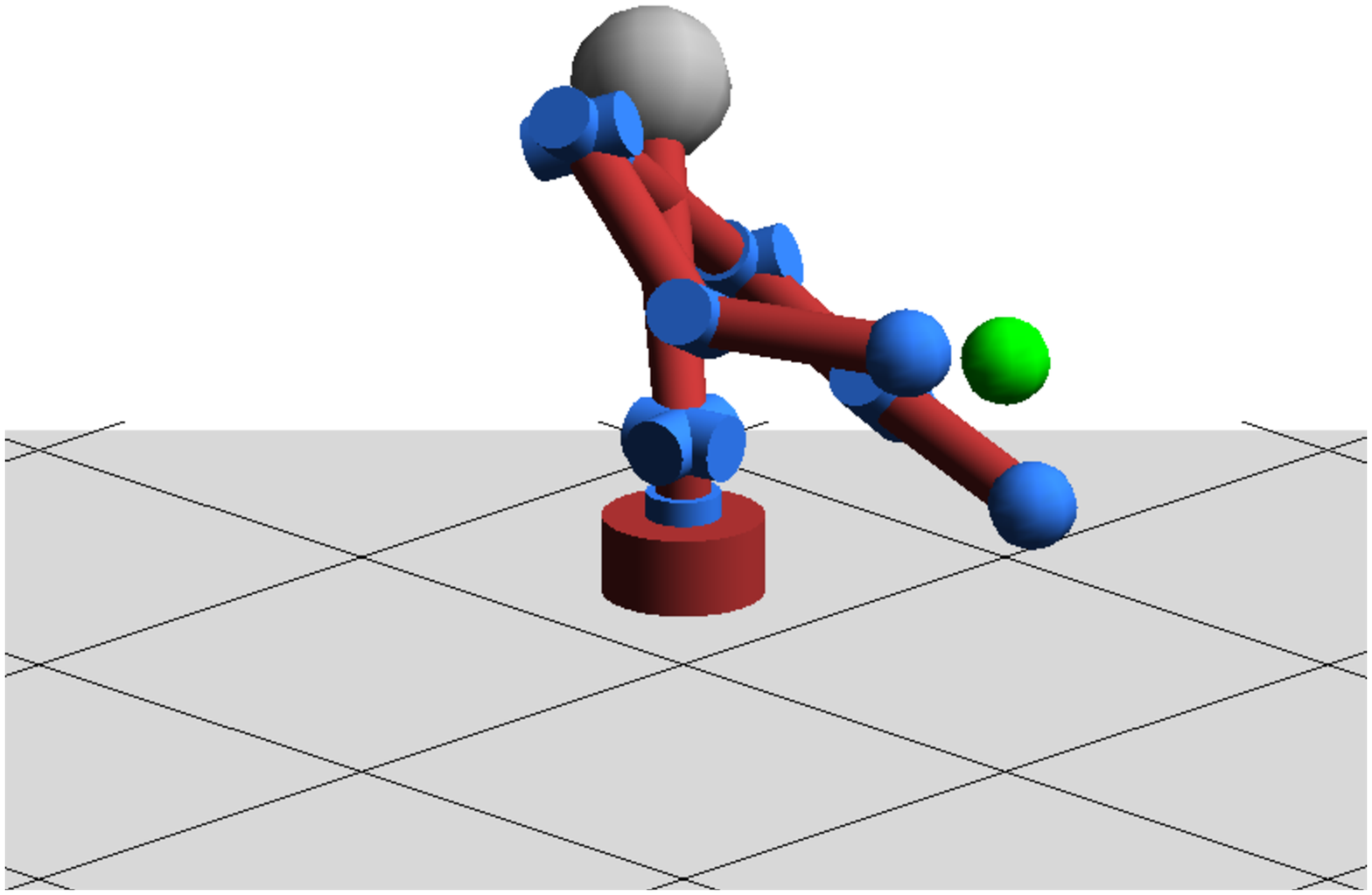}\label{fig:s13_9j}}
    \subfigure[]{\includegraphics[clip, width=0.16\columnwidth]{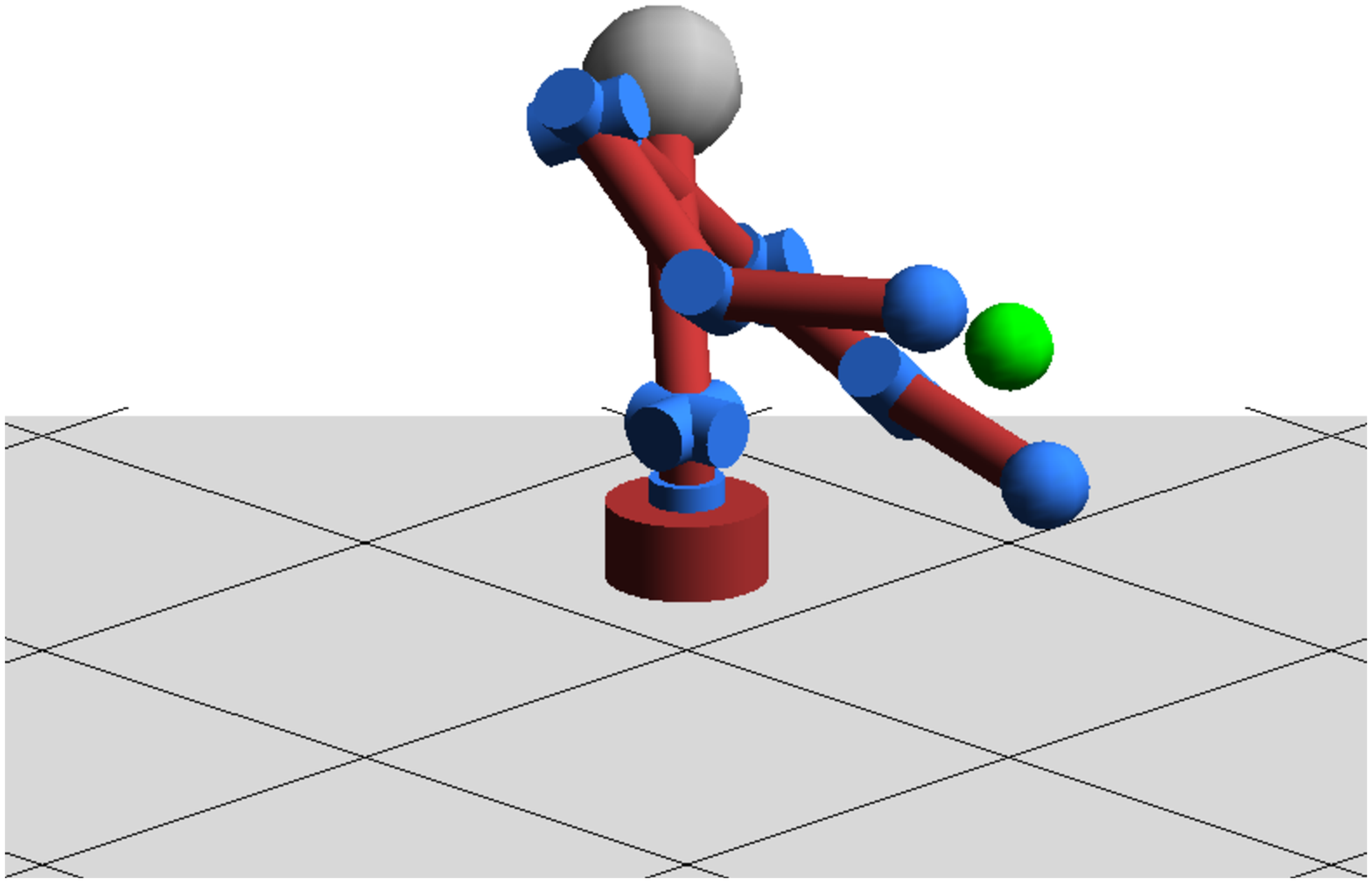}\label{fig:s14_9j}}
\caption{Example of arm reaching with $9$ joints using a policy obtained by M-PGPE(LSCDE) at the $1000$th iteration. (some intermediate steps are not shown here).}
\label{fig:simu-traject9j}
\end{figure*}

Figure~\ref{fig:simu-traject9j} shows a typical example of the reaching motion with 9 joints obtained by M-PGPE(LSCDE) at the 1000th iteration. The images show that the policy learned by M-PGPE(LSCDE) leads the right hand to the distant object successfully within 14 steps.

Overall, the proposed M-PGPE(LSCDE) method is shown to be promising in the noisy and high-dimensional humanoid robot arm reaching task.

\section{Conclusion}
\label{sec:conclusion}

We extended the model-free PGPE method to a model-based scenario,
and proposed to combine it with a model estimator called LSCDE.
Under the fixed sampling budget,
appropriately designing a sampling schedule is critical
for the model-free IW-PGPE method,
while this is not a problem for the proposed model-based PGPE method.
Through experiments, we confirmed that GP-based model estimation is not as flexible as
the LSCDE-based method when the transition model is not Gaussian,
and the proposed model-based PGPE based on LSCDE was overall demonstrated to be promising.

\section*{Acknowledgments}

VT was supported by the JASSO scholarship,
TZ was supported by the MEXT scholarship,
JM was supported by MEXT KAKENHI 23120004,
and MS was supported by the FIRST project.

\bibliographystyle{plain}

\end{document}